\documentclass{article} %

\PassOptionsToPackage{table}{xcolor}

\usepackage[preprint]{iclr2027_conference}

\usepackage[utf8]{inputenc} %
\usepackage[T1]{fontenc}    %
\usepackage{float}
\usepackage{hyperref}       %
\usepackage{url}            %
\usepackage{booktabs}       %
\usepackage{amsfonts}       %
\usepackage{nicefrac}       %
\usepackage{microtype}      %
\usepackage[table]{xcolor}         %
\usepackage{graphicx}       %

\usepackage{algorithm}
\usepackage{algpseudocode}

\usepackage{multirow}
\usepackage{tabularx}
\usepackage{newtxtext, newtxmath} %
\usepackage{pifont} %
\usepackage{makecell}
\usepackage{enumitem}
\usepackage{placeins} %
\setlist[itemize]{leftmargin=3.5mm}

\definecolor{cadmiumgreen}{rgb}{0.0, 0.42, 0.24}
\definecolor{citecolor}{HTML}{0071bc}
\definecolor{linkred}{HTML}{B22222}

\hypersetup{
  pdftitle={Why, Where, How: Taxonomy-guided Error Grounding for Code Repair in NL2SQL},
  pdfauthor={Suchan Lee, Woomin Song, Hwanjo Yu, Sangwoo Mo},
  urlcolor = cadmiumgreen,
  linkcolor = linkred,
    citecolor = cadmiumgreen,
  colorlinks = true,
}

\usepackage{xspace}

\usepackage{etoc}
\usepackage{minitoc}
\etocsettocstyle{\section*{Appendix}}{}
\newcommand{\llname}{\textbf{T}axonomy-guided \textbf{E}rror \textbf{G}rounding\xspace}

\newcommand{\sname}{TEG\xspace}

\colorlet{tabtint}{gray!10}

\newcommand{\family}[2]{\rowcolor{tabtint}\multicolumn{#1}{l}{\textbf{#2}}}

\title{\raggedright 
Why, Where, How: Taxonomy-guided Error Grounding for Code Repair in NL2SQL}

\author{%
Suchan Lee$^{1}$ \quad
Woomin Song$^{2}$ \quad
Hwanjo Yu$^{1, \dagger}$ \quad
Sangwoo Mo$^{1, \dagger}$ \\
$^{1}$POSTECH \quad
$^{2}$KAIST \\
{\normalfont\small
$^{1}$\texttt{\{leesc0324,hwanjoyu,sangwoo.mo\}@postech.ac.kr} \quad
$^{2}$\texttt{woomin.song@kaist.ac.kr}}
}

\begin{document}
\maketitle

\etocdepthtag.toc{mtchapter}
\etocsettagdepth{mtchapter}{subsection}
\etocsettagdepth{mtappendix}{none}
\faketableofcontents

\begin{abstract}
SQL queries that large language models write from natural language questions can execute successfully yet produce incorrect results, so execution alone does not reveal what to fix.
An error taxonomy says \emph{why} the query is wrong, but not \emph{where} to look or \emph{how} to change it.
Existing methods can guide SQL correction through feedback, error reports, or generated plans alongside an unmasked query.
We introduce \textbf{\sname} (\llname), which turns a supplied diagnosis into a structured correction input for natural language-to-SQL (NL2SQL) correction.
Type-specific rules map each error type to construct classes to reconsider and an edit operation to request.
TEG masks the selected constructs in the query when applicable and states that operation in an edit instruction.
TEG generates candidate corrections from this input, uses execution feedback to guide candidate selection, and repeats the process one annotation at a time for queries with several errors.
On NL2SQL-BUGs, TEG reaches 47.3 single-error execution accuracy and 37.0 overall with Qwen2.5-7B-Instruct.
Across the model sizes and thinking modes evaluated in the main comparison, TEG outperforms all evaluated baselines on single-error queries, even when the baselines receive the same error-type annotations.
With predicted types, TEG stays above direct LLM correction and ErrorLLM on single-error queries.
\end{abstract}

\begingroup
\renewcommand{\thefootnote}{\fnsymbol{footnote}}
\footnotetext[2]{Equal advising. Code: \url{https://github.com/isuchan0212/why-where-how}}
\endgroup

\section{Introduction}
Large language models (LLMs) now generate code from natural language specifications~\citep{chen2021evaluating, jiang2026survey}.
A generated program can be syntactically valid and run without an exception while still returning a wrong result, so execution alone does not establish correctness and semantic errors must be repaired~\citep{liu2023your, pan2023logic, chen2024deep, li2026bridging}.
Natural language-to-SQL (NL2SQL) is an acute case: agents increasingly rely on generated SQL over large relational databases~\citep{lei2025spider}, and a wrong query can return a plausible table rather than an error~\citep{liu2025nl2sql}.
A diagnosis from an error taxonomy tells the LLM \emph{why} the query is wrong, but not \emph{where} to look in the query or \emph{how} to change it.

Reasoning and self-correction methods~\citep{kojima2022large, madaan2023self, chen2024teaching} supply no error-type diagnosis, so the model must discover the why, the where, and the how from the code and available feedback.
NL2SQL pipelines such as DIN-SQL, MAC-SQL, and SQLFixAgent combine SQL generation with self-correction or execution-guided refinement~\citep{pourreza2023din, wang2025mac, cen2025sqlfixagent}.
ErrorLLM~\citep{hong2026errorllm} is a closely related approach: it starts from an error type, locates candidate syntax-tree nodes and schema elements under type-specific guidelines, and supplies them as context next to the issued query for a single refinement call.
TEG turns the supplied diagnosis into a correction input: deterministic rules select construct classes to mask when applicable and specify the requested edit.
\begin{figure}[t]
\centering
\includegraphics[width=\linewidth]{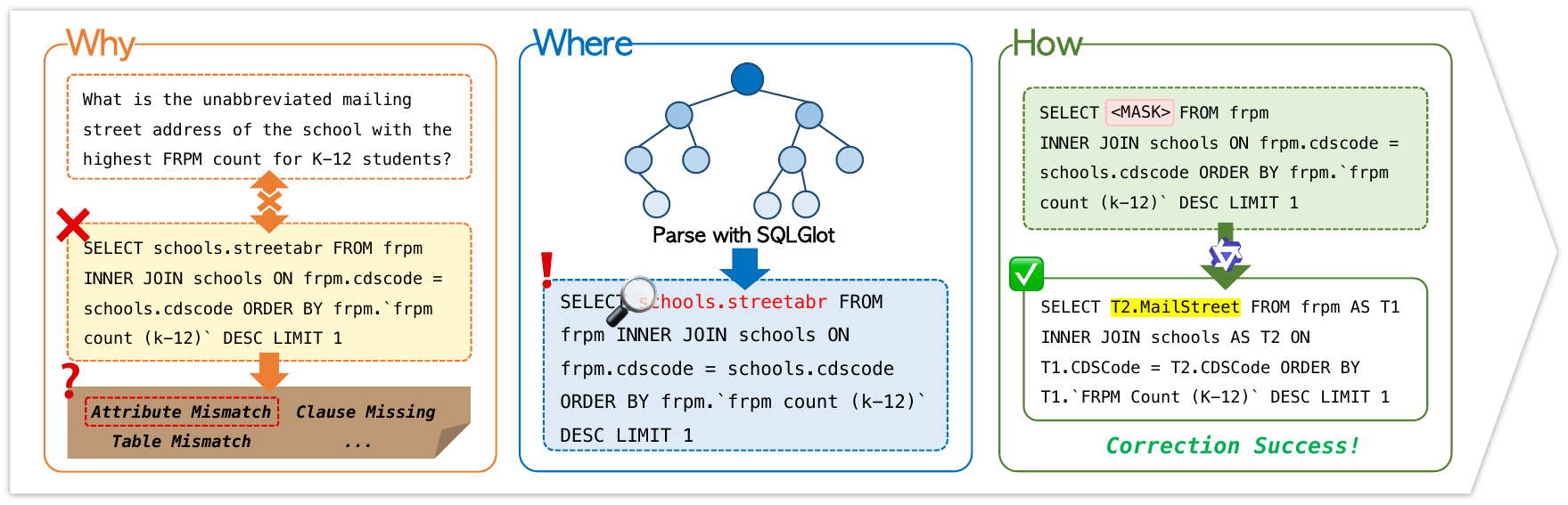}
\vspace{-7mm} %
\caption{\textbf{Error grounding maps \emph{why} a query is wrong to \emph{where} and \emph{how} to fix it.} TEG masks the constructs selected by the error type and supplies an edit instruction. For an \textit{Attribute Mismatch}, TEG masks column expressions, allowing the LLM to replace the abbreviated street field \texttt{schools.streetabr} with the required unabbreviated mailing street field \texttt{T2.MailStreet}.}
\label{fig:comparison}
\vspace{-2.5mm} %
\end{figure}

We call this mapping \emph{error grounding} (Figure~\ref{fig:comparison}): translating a taxonomy-level diagnosis into the classes of query constructs to reconsider and the edit operation to request.
For example, a diagnosis of an incorrect reference selects the class of expressions that carry references and requests a replacement, whereas a missing construct selects no existing element and requests an insertion.
A syntax tree makes these classes addressable, so the selection can be applied to the query directly instead of predicting the exact faulty occurrence.
SQL offers both an error taxonomy and structural editing rules: it exposes recurring constructs over an explicit schema, and a SQL error taxonomy labels semantic errors directly in terms of these constructs~\citep{liu2025nl2sql}.

We propose \textbf{\sname} (\llname), which realizes error grounding for SQL on the syntax tree that SQLGlot parses from the query.
The \emph{where} rule maps each supplied error type to the constructs to reconsider: TEG masks the associated syntax-tree nodes when applicable, such as column expressions for \textit{Attribute Mismatch}.
The \emph{how} rule maps the same type to an edit instruction, such as replacing the masked construct for a \textit{Mismatch} or inserting an absent one for a \textit{Missing} error.
Because the rules select construct classes, masks can cover correct constructs; the LLM generates a complete query from the grounded input.

Grounding sets where and how, but in our development runs a single output often named a column or value not in the database.
TEG therefore generates several candidates, prefers those that execute and return rows, and lets an LLM refiner select or rewrite among them, a stage we develop from SQLFixAgent~\citep{cen2025sqlfixagent} for grounded prompts.
With several errors, one prompt carrying every error type would mix several masks and instructions. TEG instead processes one annotation at a time, editing existing constructs before inserting new ones, and re-grounds the output of each step.

We evaluate TEG on incorrect queries from NL2SQL-BUGs~\citep{liu2025nl2sql}, derived from BIRD~\citep{li2023can}, against the methods above and our adaptation of ErrorLLM.
TEG outperforms every evaluated baseline on single-error queries across the model sizes and thinking modes evaluated in the main comparison, and the lead holds when baselines receive the same annotations.
On Qwen2.5-7B, it reaches 47.3 execution accuracy on single-error queries and 37.0 over all queries, against 33.6 and 24.0 for ErrorLLM, and leads every baseline on multi-error queries.
With predicted types, TEG stays above direct LLM correction and ErrorLLM on single-error queries across all evaluated configurations.
The w/o masking variant retains the error types but reduces single-error accuracy from 47.3 to 40.9, supporting the masking-based editing configuration within TEG.

In summary, our main contributions are:

\vspace{-2mm} %
\begin{itemize}

    \item \textbf{We formulate error grounding for SQL correction.} We map a supplied diagnosis (\emph{why}) to construct classes to reconsider (\emph{where}) and an edit instruction (\emph{how}). This mapping constructs the correction input without requiring exact fault localization.

    \item \textbf{We design type-specific rules and sequential re-grounding in \sname.} Four edit groups cover the 29 evaluated NL2SQL-BUGs sub-types, with masking when applicable and type-specific edit instructions. For multiple errors, TEG reapplies the rules to the updated query for each annotation.

    \item \textbf{We demonstrate improved SQL correction with TEG.} TEG achieves the highest observed accuracy in the main comparisons, including those with shared annotations. We also evaluate predicted-type correction on single-error queries and compare component variants to examine grounding, sequential correction, and candidate selection within the pipeline.
\end{itemize}

\section{Related Work}
\textbf{Reasoning and self-correction.}
Reasoning prompts such as CoT~\citep{kojima2022large, wei2022chain} elicit step-by-step analysis, while Self-Refine~\citep{madaan2023self} and Self-Debug~\citep{chen2024teaching} guide correction with natural language critiques, code explanations, or execution feedback.
Self-correction can also be learned~\citep{kumar2025training}, although improvement through intrinsic feedback alone is not guaranteed~\citep{huang2024large}.
These methods supply no error-type diagnosis, so the model must decide why, where, and how to correct from the code and its feedback alone; TEG instead assumes the diagnosis is given and asks how best to use it.

\textbf{SQL-specific refinement.}
LLMs have enabled diverse NL2SQL approaches~\citep{liu2025survey}.
These span prompt design, task-specific training, structure-guided generation, and test-time scaling~\citep{dong2023c3, li2024codes, zhang2024structure, liao2026learnat, he2025star, li2025alpha, wang2025agentar}.
Refinement and debugging pipelines use self-correction, execution feedback, or alternative query representations~\citep{talaei2024chess, askari2025magic, xu2025ts, li2026swe, qu2025share}.
DIN-SQL~\citep{pourreza2023din} couples decomposed generation with self-correction, MAC-SQL~\citep{wang2025mac} coordinates schema selection, decomposition, and execution-guided refinement, and SQLFixAgent~\citep{cen2025sqlfixagent} separates query review, candidate generation, and refinement across agents.
Candidate validation and selection are also studied~\citep{fan2025grounding, gorti2025msc, lee2025mcs, pourreza2025chase}.
TEG develops its execution-based candidate selection from SQLFixAgent~\citep{cen2025sqlfixagent} for grounded inputs.

\textbf{Program repair and code infilling.}
DrRepair localizes faulty lines~\citep{yasunaga2020graph}, while LecPrompt uses CodeBERT for token-level localization and repair~\citep{xu2024lecprompt, feng2020codebert}.
Masked-span reconstruction~\citep{guo2022unixcoder} and code infilling~\citep{fried2022incoder, allal2023santacoder} recover missing code from surrounding context.
TEG selects construct classes by error type without exact fault localization and uses masks when applicable and edit instructions as correction inputs. It trains no infilling model; the LLM generates full queries and may edit unmasked constructs.

\textbf{Error taxonomies and detection.}
Error taxonomies give diagnoses reusable categories, from software defect classifications~\citep{chillarege1992orthogonal} to the semantic SQL errors annotated in NL2SQL-BUGs~\citep{liu2025nl2sql}.
Related SQL error detection methods assess generated queries using learned structural representations~\citep{chen2023error} or metamorphic testing~\citep{yang2025hallucination}.
Multi-grained error identification~\citep{xu2025boosting} distinguishes system, skeleton, and value errors and uses type-specific feedback to guide SQL correction.
SQLens~\citep{gong2026sqlens} combines database and LLM signals into clause-level error reports with correction instructions and uses these reports to guide sequential refinement of the generated query.

\textbf{Taxonomy-guided refinement.}
SQL-of-Thought~\citep{chaturvedi2025sql} uses an error taxonomy to guide an LLM-generated correction plan, which a separate agent uses to revise the query.
ErrorLLM~\citep{hong2026errorllm} is closely related: its error taxonomy draws on NL2SQL-BUGs, and its refinement stage uses error types to guide structural correction. It trains an error detector that predicts taxonomy categories with dedicated error tokens from the question, schema, and SQL structure, and localizes relevant syntax-tree nodes and schema elements under type-specific guidelines.
The error type also governs the scope of context extraction, and the resulting subtrees, schema context, and analyses are supplied alongside the issued query for joint refinement.
We adapt its refinement stage as a principal baseline, using the same supplied error annotations as TEG.
TEG maps the supplied diagnosis (\emph{why}) through deterministic rules to construct-class masks (\emph{where}) and edit instructions (\emph{how}), removing selected constructs from the query input when masking applies.

\section{TEG: Taxonomy-guided Error Grounding}
\label{sec:method}
\begin{figure}[t]
\centering
\includegraphics[width=\linewidth]{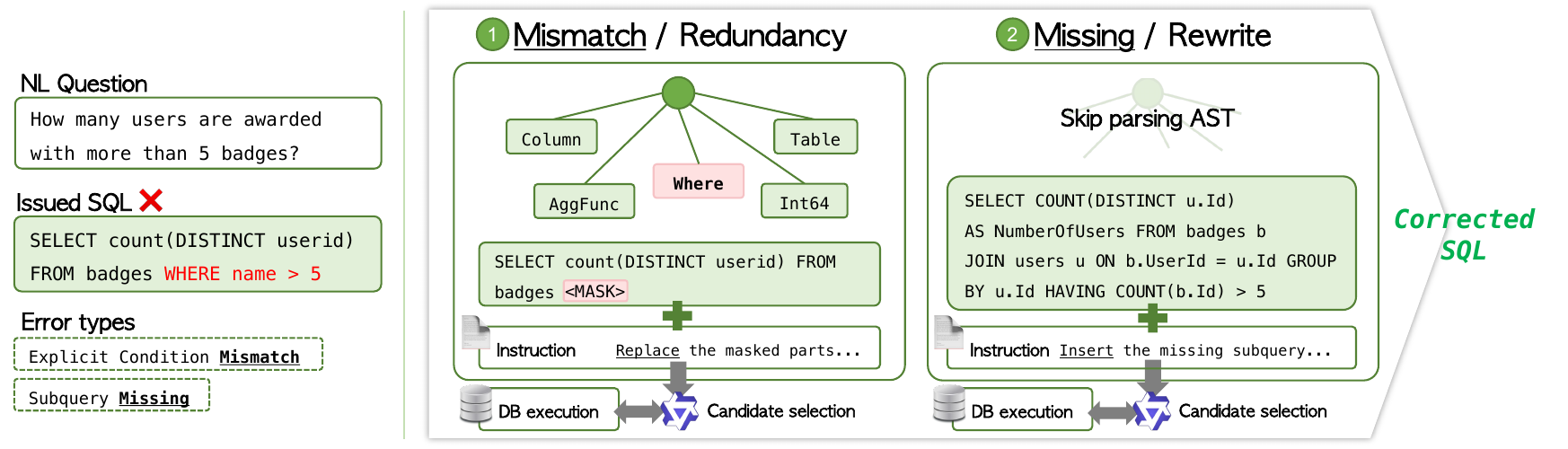}
\vspace{-6mm} %
\caption{\textbf{TEG maps error types to masking rules and edit instructions.} The \emph{where} rule masks constructs for \textit{Mismatch} and \textit{Redundancy} (1); \textit{Missing} and \textit{Rewrite} (2) leave the query unmasked. The \emph{how} rule specifies the edit. Here, step~1 replaces the masked \texttt{WHERE} predicate for an \textit{Explicit Condition Mismatch}; step~2 inserts the missing subquery for \textit{Subquery Missing} into step~1\textquotesingle s output.}
\label{fig:methodology}
\vspace{-3mm} %
\end{figure}

TEG turns error annotations into grounded SQL correction steps (Figure~\ref{fig:methodology}, Algorithm~\ref{alg:step}).
Each step masks the constructs that the error type selects in the SQLGlot syntax tree, when there are any, states the requested edit, and selects the corrected query from candidates by execution; a query with several annotations takes one step per annotation in a fixed order.

Appendix~\ref{app:implementation} gives the full rules, masking algorithms, prompts, and candidate selection details.

\subsection{Grounding error types into masks and edit instructions}
\label{sec:formulation}

Grounding turns an error type, which says \emph{why} a query is wrong, into two parts of the correction input rather than of its output: a mask over the constructs to reconsider, which tells the LLM \emph{where} to look, and an edit instruction, which tells it \emph{how} to change them.
The mask removes the selected constructs from the query the LLM sees, and the edit instruction names the requested edit, so one error type determines both what the LLM reconsiders and what it is asked to do.

TEG prompts an off-the-shelf LLM with an issued query, its question, schema, and error annotations, and returns a corrected query.
Let $q$ be the issued query, $n$ its natural language question, and $S$ the database schema.
We assume an annotation sequence $\mathcal{E}=((t_1,s_1),\ldots,(t_m,s_m))$, where each pair specifies an error category (e.g., \textit{Attribute-Related Errors}) and sub-type (e.g., \textit{Attribute Mismatch}).
The goal is a corrected SQL query $q^*$ semantically aligned with $n$ under schema $S$.
We call $|\mathcal{E}|=1$ the single-error setting and $|\mathcal{E}|>1$ the multi-error setting.

The grounding rules act on four edit groups that TEG defines over the error taxonomy of NL2SQL-BUGs~\citep{liu2025nl2sql}. The queries we evaluate cover nine categories and 29 observed sub-types, classified by the SQL construct they affect, such as \textit{Attribute-Related}, \textit{Table-Related}, and \textit{Clause-Related}.
The four groups are \textit{Mismatch}, \textit{Redundancy}, \textit{Missing}, and \textit{Rewrite}, and the mapping from sub-types to groups is part of TEG and specifies the default edit operation and correction order. The annotations retain their original taxonomy categories and sub-types.

The \emph{where} rule maps the category and sub-type to classes of nodes in the abstract syntax tree (AST) to mask. For \textit{Mismatch} and \textit{Redundancy}, TEG masks matching existing constructs. \textit{Missing} requests an absent construct and \textit{Rewrite} requests broader restructuring, so TEG leaves the query unmasked for both. For example, \textit{Attribute Mismatch} hides column expressions, \textit{Explicit Condition Mismatch} hides the \texttt{WHERE} predicate, and \textit{Aggregate Functions} hides the aggregate together with the grouping and filtering clauses that depend on it.
The rules also set the granularity of masking: hiding a single node can leave a dependent clause inconsistent after repair, whereas hiding a whole clause can discard context the LLM would otherwise keep, so each rule sets the unit per error type.

The \emph{how} rule specifies an edit instruction for the supplied error type: replace for \textit{Mismatch}, remove for \textit{Redundancy}, insert for \textit{Missing}, and restructure for \textit{Rewrite}, which covers incomplete queries and errors that need broader changes. Sub-types such as \textit{Aggregate Functions} and \textit{ASC/DESC} are grouped with \textit{Mismatch} for the correction order, but their edit instructions depend on whether a mask is present: replacement or deletion with a mask, and insertion without one. For the \textit{DISTINCT} sub-type, the instruction also permits restructuring joins, projection, or grouping.

\begin{algorithm}[t]
\caption{\textbf{TEG re-grounds the current query for each error annotation.} $\textsc{TEG}(q,\, n,\, S,\, \mathcal{E})$ processes the nonempty sequence $\mathcal{E}$ in edit-group order and applies candidate selection at each step; the retry branch of Section~\ref{sec:multi-error} is omitted. Each updated query becomes the input to the next step.}
\label{alg:step}
\begin{algorithmic}[1]
\Require SQL $q$, question $n$, schema $S$, annotation sequence $\mathcal{E}$, $|\mathcal{E}|>0$
\Ensure Corrected SQL $q^*$
\State $N \leftarrow \{n\} \cup \textsc{Craft}(n)$ \Comment{question paraphrases, produced once}
\State Stable-sort $\mathcal{E}$ by edit group (Mismatch, Redundancy, Missing, Rewrite)
\For{each $(t,s)$ in $\mathcal{E}$, in order}
    \State $\tilde{q} \leftarrow \textsc{Masking}(q,\, t,\, s)$ \Comment{type-specific masking; no mask for Missing or Rewrite}
    \State $C \leftarrow \{\textsc{LLM}(\textsc{Prompt}(\tilde{q},\, n',\, S,\, t,\, s)) : n' \in N\}$ \Comment{one candidate per question variant}
    \State $q \leftarrow \textsc{Select}(C,\, n,\, S,\, q)$ \Comment{execution filter and refiner}
\EndFor
\State \Return $q$
\end{algorithmic}
\end{algorithm}

\subsection{Grounded correction with candidate selection}
\label{sec:step}

A correction step applies both rules to one annotation and asks the LLM for a repaired query. For \textit{Mismatch} and \textit{Redundancy}, TEG parses the current query with SQLGlot and replaces the selected constructs with \texttt{<MASK>} placeholders; \textit{Missing} and \textit{Rewrite} leave the query unmasked.
Attribute masking locates column expressions in the parsed tree and replaces their SQL strings, whereas the other masking rules replace the AST nodes themselves.
If parsing or masking fails, the query is passed unchanged to the LLM.
The prompt includes the possibly masked query, together with the question, the schema, and the edit instruction. When masking is applied, the prompt does not include the unmasked query, so the LLM resolves each placeholder without seeing the removed constructs.

Masking removes the selected constructs from the input, whereas naming the error type leaves them in place. The rules select node classes deterministically, but the LLM decides what to change: several nodes may match, and some may already be correct. The LLM generates a complete query, so the output can revise unmasked constructs as well as fill the placeholders.

TEG selects the corrected query from several candidates with execution feedback, because one answer to a grounded prompt can name a column or value absent from the database.
We develop this selection stage from the QueryCrafter and SQLRefiner of SQLFixAgent~\citep{cen2025sqlfixagent} for grounded prompts, combining candidate generation, execution filtering, and LLM-based selection or refinement.
The original question and up to three paraphrases yield at most four candidates per step. The paraphrases are generated once and reused as later steps correct the updated query.

\subsection{Sequential re-grounding for multiple errors}
\label{sec:multi-error}

We design sequential correction to process one annotation at a time and re-ground the query after each update, so that each step targets one error type with its own mask and instruction.

TEG uses the order \textit{Mismatch}, \textit{Redundancy}, \textit{Missing}, \textit{Rewrite}, preserving input order within groups.
We edit existing constructs first because an insertion or restructuring can move or remove the constructs a later rule would select.
Re-grounding applies the current annotation to the query produced by earlier steps, masking it only when applicable. TEG retains the supplied annotations without predicting new error types, and later edits can undo an earlier repair. If masking yields no placeholder and a previous correction step exists, TEG uses the current query without masking, together with a retry instruction that names both the current and the previous errors.
If $\mathcal{E}$ is empty, as when no error type is predicted, TEG falls back to one unmasked correction call and keeps $q$ if it yields no SQL.

\section{Experiments}
\begin{table}[t]
\centering\small
\caption{
\textbf{TEG leads single-error correction across the evaluated models.}
Entries are execution accuracies (\%) on single-error queries from NL2SQL-BUGs.
In the Error types column, \ding{51} denotes supplied oracle error types from NL2SQL-BUGs; \ding{55} denotes no error types. ErrorLLM denotes our adaptation (Section~\ref{sec:setup}). Qwen3-8B is evaluated with thinking OFF and ON. Bold marks the best accuracy per Qwen setting.
TEG\textquotesingle s margin over the strongest baseline ranges from 0.9 to 11.8 points.
}
\label{tab:baseline_single}
\setlength{\tabcolsep}{3.8pt}
\begin{tabular}{l|c|ccc|cc}
\toprule
& & \multicolumn{3}{c|}{\textbf{Qwen2.5-Instruct}} & \multicolumn{2}{c}{\textbf{Qwen3-8B}} \\
Method & Error types & 7B & 14B & 32B & Thinking OFF & Thinking ON \\
\midrule
Base & \ding{55} & 30.0 & 35.5 & 41.8 & 31.8 & 43.6 \\
\midrule \addlinespace[0.3em]
\family{7}{Reasoning and self-correction} \\
\addlinespace[0.3em]
CoT~\citep{kojima2022large} & \ding{55} & 35.5 & 37.3 & 45.5 & 33.6 & 48.2 \\
Self-Debug (Simple)~\citep{chen2024teaching} & \ding{55} & 33.6 & 38.2 & 48.2 & 38.2 & 50.9 \\
Self-Debug (Expl)~\citep{chen2024teaching} & \ding{55} & 29.1 & 35.5 & 44.5 & 44.5 & 47.3 \\
\midrule \addlinespace[0.3em]
\family{7}{SQL-specific refinement} \\
\addlinespace[0.3em]
DIN-SQL~\citep{pourreza2023din} & \ding{55} & 24.5 & 42.7 & 53.6 & 21.8 & 43.6 \\
MAC-SQL~\citep{wang2025mac} & \ding{55} & 18.2 & 15.5 & 18.2 & 25.5 & 30.0 \\
SQLFixAgent~\citep{cen2025sqlfixagent} & \ding{55} & 32.7 & 36.4 & 43.6 & 37.3 & 42.7 \\
\midrule \addlinespace[0.3em]
\family{7}{Taxonomy-guided refinement} \\
\addlinespace[0.3em]
ErrorLLM~\citep{hong2026errorllm} & \ding{51} & 33.6 & 51.8 & 53.6 & 39.1 & 45.5 \\
\midrule \addlinespace[0.3em]
\family{7}{Taxonomy-guided error grounding} \\
\addlinespace[0.3em]
\sname (ours) & \ding{51} & \textbf{47.3} & \textbf{52.7} & \textbf{57.3} & \textbf{46.4} & \textbf{57.3} \\
\bottomrule
\end{tabular}
\end{table}

Our experiments ask three questions about error grounding, in the order a reader would raise them:
\begin{itemize}
    \item \textbf{How does TEG compare with existing correction methods?} We compare TEG with correction baselines, including versions given the same error types (Section~\ref{sec:main}).
    \item \textbf{Does the gain survive predicted error types?} We replace the benchmark's error annotations with error types predicted by Opus~4.8 and ask how much of the lead remains (Section~\ref{sec:predicted}).
    \item \textbf{How do component variants compare, and what is the computational cost?} We ablate the mask, the sequential steps, and the candidate selection, compare error categories, and measure model-call time, excluding execution and preprocessing (Section~\ref{sec:analysis}).
\end{itemize}

\subsection{Setup}
\label{sec:setup}

We evaluate on NL2SQL-BUGs~\citep{liu2025nl2sql}, derived from BIRD~\citep{li2023can}, using incorrect queries paired with reference SQL; the evaluation subset covers 29 error sub-types across nine categories.
Execution accuracy measures whether the corrected query and the reference query return the same set of row tuples on the same database, ignoring row order and duplicates while preserving column order.
Queries that raise execution errors or exceed the 5s time limit are counted as incorrect, and all accuracy scores are reported on a 0--100 scale.
During correction, TEG executes its own candidates with a separate 5s limit and never sees the reference SQL or its result.

For the main comparisons, we use Qwen2.5-Instruct~\citep{qwen2.5} at 7B, 14B, and 32B and Qwen3-8B~\citep{yang2025qwen3} with thinking OFF and ON, all with greedy decoding, as the LLM for every method.
With each model, we compare TEG against the LLM itself and three baseline groups:
\begin{itemize}
    \item \textbf{Base}: the LLM itself, asked once to correct the unmasked query with no error type, as the reference for what each correction method adds beyond a single call.
    \item \textbf{Reasoning and self-correction}: CoT~\citep{kojima2022large} and Self-Debug in its Simple and Expl variants~\citep{chen2024teaching}, which we adapt to SQL correction by prompting them to analyze or critique the given erroneous query and produce a corrected query.
    \item \textbf{SQL-specific refinement}: DIN-SQL~\citep{pourreza2023din}, MAC-SQL~\citep{wang2025mac}, and SQLFixAgent~\citep{cen2025sqlfixagent}, which we adapt to SQL correction by providing the given erroneous query as input or context, without an error-type diagnosis.
    \item \textbf{Taxonomy-guided refinement}: our adaptation of ErrorLLM~\citep{hong2026errorllm}, which uses AST-node localization, error-specific subtree and schema-context extraction, and joint refinement. It uses the supplied annotations instead of its trained error detector, and database metadata with lexical matches to the question instead of the learned question--schema encoder.
\end{itemize}
We additionally evaluate Base, ErrorLLM, and TEG with Opus~4.8 as the corrector in Section~\ref{sec:predicted}.
\begin{table}[t]
\centering\small
\caption{
\textbf{TEG leads every error-count group with Qwen2.5-7B-Instruct.}
Entries are execution accuracies (\%) on NL2SQL-BUGs, grouped by error count; Overall covers all queries.
In the Error types column, \ding{51} denotes supplied oracle error types from NL2SQL-BUGs; \ding{55} denotes no error types. ErrorLLM denotes our adaptation (Section~\ref{sec:setup}). Bold marks the best value per column.
TEG retains its lead on multi-error queries, although its accuracy is lower for queries with more errors.
}
\label{tab:nl2sql_results_multi}
\setlength{\tabcolsep}{4pt}
\begin{tabular}{l|c|ccc|c}
\toprule
Method & Error types & 1 error & 2 errors & 3+ errors & Overall \\
\midrule
Base & \ding{55} & 30.0 & 25.9 & 14.1 & 23.7 \\
\midrule \addlinespace[0.3em]
\family{6}{Reasoning and self-correction} \\
\addlinespace[0.3em]
CoT~\citep{kojima2022large} & \ding{55} & 35.5 & 21.5 & 21.8 & 24.4 \\
Self-Debug (Simple)~\citep{chen2024teaching} & \ding{55} & 33.6 & 20.9 & 16.9 & 22.4 \\
Self-Debug (Expl)~\citep{chen2024teaching} & \ding{55} & 29.1 & 24.6 & 18.3 & 23.9 \\
\midrule \addlinespace[0.3em]
\family{6}{SQL-specific refinement} \\
\addlinespace[0.3em]
DIN-SQL~\citep{pourreza2023din} & \ding{55} & 24.5 & 20.2 & 15.5 & 19.9 \\
MAC-SQL~\citep{wang2025mac} & \ding{55} & 18.2 & 14.1 & \phantom{0}8.5 & 13.5 \\
SQLFixAgent~\citep{cen2025sqlfixagent} & \ding{55} & 32.7 & 26.9 & 19.7 & 26.2 \\
\midrule \addlinespace[0.3em]
\family{6}{Taxonomy-guided refinement} \\
\addlinespace[0.3em]
ErrorLLM~\citep{hong2026errorllm} & \ding{51} & 33.6 & 23.6 & 17.6 & 24.0 \\
\midrule \addlinespace[0.3em]
\family{6}{Taxonomy-guided error grounding} \\
\addlinespace[0.3em]
\sname (ours) & \ding{51} & \textbf{47.3} & \textbf{36.7} & \textbf{29.6} & \textbf{37.0} \\
\bottomrule
\end{tabular}
\end{table}

Unless otherwise stated, supplied error types are oracle annotations from NL2SQL-BUGs. ErrorLLM and TEG receive these annotations in the main comparisons, while Base and the other six baselines receive no error types. We additionally evaluate all baselines with the same annotations.
Appendix~\ref{app:setup} gives dataset statistics, thinking modes, baseline and ablation implementations, and hardware.

\subsection{Main results}
\label{sec:main}\label{sec:single}\label{sec:multi}\label{sec:annotated}

\noindent\textbf{Single-error queries.}
Each query tests one grounded correction step without sequential re-grounding.
TEG has the highest single-error accuracy in every configuration reported in Table~\ref{tab:baseline_single}.
At 7B, it exceeds the strongest baseline, CoT, by 11.8 points (47.3 versus 35.5). Across the Qwen2.5 scales, it reaches 47.3, 52.7, and 57.3, against 33.6, 51.8, and 53.6 for ErrorLLM. On Qwen3-8B, it reaches 46.4 with thinking OFF and 57.3 with thinking ON, above the strongest baseline in each mode, Self-Debug (Expl) at 44.5 and Self-Debug (Simple) at 50.9. TEG scores above CoT at every Qwen2.5 scale and above Base in both Qwen3-8B thinking modes. On Qwen3-8B, thinking lifts Base from 31.8 to 43.6 and TEG from 46.4 to 57.3. TEG can complement model reasoning.

\noindent\textbf{Multi-error queries.}
The lead extends to queries with several errors (Table~\ref{tab:nl2sql_results_multi}).
With Qwen2.5-7B-Instruct on all queries, TEG reaches 37.0 overall against 26.2 for SQLFixAgent, the strongest baseline, and 24.0 for ErrorLLM, and it leads every error-count group, although its accuracy declines as the number of errors increases.
It also leads every error category on multi-error queries, by 5.9 to 15.4 points over the next best method.

\noindent\textbf{Baselines given error types.}
TEG\textquotesingle s lead persists when all methods receive the same error types.
To distinguish access to the error type from how it is used, we add the same annotations to each baseline\textquotesingle s prompts while retaining its correction procedure.
TEG stays ahead in all five configurations on single-error queries, by 0.9 points at Qwen2.5-14B and Qwen3-8B with thinking OFF and by 11.8 points at Qwen2.5-7B, and on all queries it reaches 37.0 against 26.0 for the strongest baseline given error types, Self-Debug (Simple), leading in every error-count group.
Error types help some baselines but not others: at 7B, CoT falls from 35.5 to 32.7, while Self-Debug (Simple) rises from 33.6 to 35.5.
Access to error types alone does not explain TEG\textquotesingle s advantage.
Appendix~\ref{app:main-comparison} reports these comparisons in full, together with the per-category breakdowns.
These comparisons evaluate complete correction pipelines: supplying the same annotations controls diagnostic information, but does not equalize candidate-generation and refinement procedures.

\subsection{End-to-end correction with predicted error types}
\label{sec:predicted}

\begin{table}[t]
\centering\small
\caption{
\textbf{Single-error correction with predicted error types.}
Entries are single-error execution accuracies (\%) on NL2SQL-BUGs.
Opus~4.8 supplies the same predicted types to TEG and ErrorLLM; it also corrects queries in the Opus~4.8 column. Base receives no types. Bold marks the best value per setting.
On empty predictions, TEG attempts unmasked correction; ErrorLLM retains the query.
}
\label{tab:predicted_single}
\setlength{\tabcolsep}{4pt}
\begin{tabular}{l|ccc|cc|c}
\toprule
& \multicolumn{3}{c|}{\textbf{Qwen2.5-Instruct}} & \multicolumn{2}{c|}{\textbf{Qwen3-8B}} & \textbf{Opus~4.8} \\
Method & 7B & 14B & 32B & Thinking OFF & Thinking ON & \\
\midrule
Base & 30.0 & 35.5 & 41.8 & 31.8 & 43.6 & 63.6 \\
\midrule \addlinespace[0.3em]
ErrorLLM & 32.7 & 45.5 & 47.3 & 38.2 & 44.5 & 64.5 \\
\sname (ours) & \textbf{45.5} & \textbf{48.2} & \textbf{52.7} & \textbf{45.5} & \textbf{53.6} & \textbf{69.1} \\
\bottomrule
\end{tabular}
\end{table}

We evaluate single-error correction using the same Opus~4.8 type predictions for TEG and ErrorLLM (Table~\ref{tab:predicted_single}). TEG outperforms Base and ErrorLLM in every evaluated corrector configuration. Across the five configurations in Table~\ref{tab:baseline_single}, its accuracy decreases by 0.9--4.6 points relative to the oracle-type results.
These gains hold even though the predicted types exactly match the benchmark annotations on only 52.7\% of single-error queries.

With Opus~4.8 for type prediction and correction, TEG scores 69.1, ErrorLLM 64.5, and Base 63.6.

The aggregate comparison includes different behavior on empty predictions: TEG makes an unmasked correction call, whereas ErrorLLM retains the issued query. On queries with nonempty predictions, TEG leads ErrorLLM by 1.1--12.6 points across the five locally hosted configurations; with Opus~4.8, the corresponding accuracies are 83.9 and 81.6. These results compare complete correction pipelines, including their fallback policies.
With oracle types on Opus~4.8, ErrorLLM and TEG achieve similar execution accuracy at 73.6 and 72.7, respectively. The fallback contributes substantially to the predicted-type lead at Qwen2.5-14B, Qwen3-8B with thinking OFF, and Opus~4.8. These comparisons do not isolate the effect of masking.
Appendix~\ref{app:predicted-types} adds prediction and category analyses.

\subsection{Ablation study and analysis}
\label{sec:analysis}

\begin{table}[t]
\centering\small
\setlength{\tabcolsep}{3pt}
\caption{
\textbf{Every ablation variant has lower overall accuracy than TEG.}
Entries are execution accuracies (\%) by error count on NL2SQL-BUGs with Qwen2.5-7B-Instruct; Overall covers all queries.
Bold marks the best value per column, including ties.
Without candidate selection or sequential correction, single-error accuracy ties the full method.
}
\label{tab:analysis}
\begin{tabular}{l|ccc|c}
\toprule
Method & 1 error & 2 errors & 3+ errors & Overall \\
\midrule
\sname (ours) & \textbf{47.3} & \textbf{36.7} & \textbf{29.6} & \textbf{37.0} \\
\midrule
\quad w/o error annotation & 38.2 & 34.7 & 21.1 & 31.9 \\
\quad w/o masking & 40.9 & 34.3 & 27.5 & 33.9 \\
\quad w/o DB execution & 40.9 & 27.3 & 19.0 & 27.9 \\
\quad w/o candidate selection & \textbf{47.3} & 25.9 & 20.4 & 28.8 \\
\quad w/o sequential correction & \textbf{47.3} & 32.3 & 23.9 & 33.2 \\
\bottomrule
\end{tabular}
\end{table}

\begin{table}[t]
\centering\small
\caption{
\textbf{TEG leads on Table, Condition, and Attribute errors with Qwen2.5-7B.}
Entries are single-error execution accuracies (\%) by oracle error category with Qwen2.5-7B-Instruct.
In the Error types column, \ding{51} denotes supplied oracle error types from NL2SQL-BUGs; \ding{55} denotes no error types. Others pools the less frequent categories. Bold marks the best value per column, including ties.
The largest gains over Base are on Table and Condition errors.
}
\label{tab:analysis_category}
\setlength{\tabcolsep}{4pt}
\begin{tabular}{l|c|ccccc|c}
\toprule
Method & Error types & Attribute & Value & Condition & Table & Others & Overall \\
\midrule
Base & \ding{55} & 24.3 & 41.4 & 29.4 & 33.3 & 16.7 & 30.0 \\
CoT & \ding{55} & 35.1 & 37.9 & 52.9 & 26.7 & 16.7 & 35.5 \\
Self-Debug (Simple) & \ding{55} & 24.3 & 41.4 & 35.3 & 53.3 & 16.7 & 33.6 \\
Self-Debug (Expl) & \ding{55} & 24.3 & 37.9 & 35.3 & 33.3 & \phantom{0}8.3 & 29.1 \\
DIN-SQL & \ding{55} & 10.8 & \textbf{51.7} & 23.5 & 13.3 & 16.7 & 24.5 \\
MAC-SQL & \ding{55} & 10.8 & 27.6 & 23.5 & \phantom{0}6.7 & \textbf{25.0} & 18.2 \\
SQLFixAgent & \ding{55} & 24.3 & 41.4 & 41.2 & 40.0 & 16.7 & 32.7 \\
ErrorLLM & \ding{51} & 16.2 & 44.8 & 41.2 & 60.0 & 16.7 & 33.6 \\
\midrule
\rowcolor{tabtint} \sname (ours) & \ding{51} & \textbf{37.8} & \textbf{51.7} & \textbf{58.8} & \textbf{66.7} & \textbf{25.0} & \textbf{47.3} \\
\bottomrule
\end{tabular}
\end{table}

\noindent\textbf{Ablation study.}
Table~\ref{tab:analysis} compares TEG with ablation variants using Qwen2.5-7B-Instruct.
The w/o masking variant retains the error annotations and candidate-generation and selection procedures, but leaves the query unmasked. Accuracy falls from 47.3 to 40.9 on single-error queries and from 37.0 to 33.9 overall. The variant also disables the mask-failure retry branch on multi-error queries, and mask-dependent edit instructions can change. These results support the masking-based editing configuration within TEG; they do not isolate the effect of masking the input alone.
Removing the annotation as well lowers single-error and overall accuracy to 38.2 and 31.9.
This variant also drops sequential correction and does not match compute on multi-error queries, so the rows measure interacting components rather than additive contributions.

Correcting all errors in one pass with cumulative masks, while retaining candidate selection, lowers accuracy from 36.7 to 32.3 on two errors and from 29.6 to 23.9 on three or more, for 33.2 overall. Single-error accuracy is unchanged because both variants take one correction step.
These results favor sequential re-grounding: each step applies type-specific rules to the query produced by earlier corrections. The one-pass variant lists all annotations together and uses the first annotation's category to select a single prompt template.

The w/o candidate selection variant uses one generation call per correction step without the crafter or refiner; it ties TEG at 47.3 on single-error queries but falls to 28.8 overall. The w/o DB execution variant retains multiple candidates but limits refinement to one round without execution feedback, lowering accuracy to 40.9 on single-error queries and 27.9 overall.
A single candidate per step retains TEG's single-error lead over the baselines; candidate selection helps on multi-error queries. On single-error queries, the w/o DB execution variant trails the single-candidate variant (40.9 versus 47.3), showing that generating more candidates does not by itself ensure better correction.

\noindent\textbf{Per-category results.}
TEG outperforms every baseline on Table, Condition, and Attribute errors with Qwen2.5-7B (Table~\ref{tab:analysis_category}).
The largest gains over Base are on Table errors (66.7 versus 33.3) and Condition errors (58.8 versus 29.4).
On Attribute errors, TEG reaches 37.8, compared with 24.3 for Base and 16.2 for ErrorLLM, which receives the same error types.
TEG also ties DIN-SQL on Value errors at 51.7 and MAC-SQL on Others at 25.0, so its lead varies across categories.
For maskable sub-types in these categories, the \emph{where} rule focuses masking on the relevant construct classes---table and join nodes, column expressions, or \texttt{WHERE} predicates---while the \emph{how} rule specifies the edit. This direct correspondence may help explain the gains.
For the \textit{Missing} group, TEG requests insertion without masking.
These results support the effectiveness of the full TEG pipeline in these categories, but do not isolate masking from differences in edit instructions and candidate selection.

\noindent\textbf{Computational cost.}
The multi-error gain from several candidates costs extra model computation.
With Qwen2.5-7B-Instruct, mean summed model-call time per query is 42.9s for TEG, 8.6s for the single-candidate variant, and 3.3s for Base, excluding execution and preprocessing.
Compared with the single-candidate variant, TEG gains 10.8 points on two errors and 9.2 on three or more, while the two methods have the same accuracy on single-error queries.

\section{Conclusion and Discussion}
\label{sec:conclusion}\label{sec:discussion}\label{sec:limitations}
We introduced error grounding for SQL correction: mapping \emph{why} a query is wrong to construct classes (\emph{where}) and edit instructions (\emph{how}) that shape the LLM\textquotesingle s correction input.
On NL2SQL-BUGs, TEG leads the main comparisons even with shared annotations and outperforms direct correction and ErrorLLM with predicted types on single-error queries. The w/o masking variant retains the types but lowers single-error accuracy by 6.4 points with Qwen2.5-7B. These results support using error types to construct correction inputs through masking and edit instructions.

\noindent\textbf{Beyond SQL.}
Extending error grounding beyond SQL would retain the mapping from error types to editable structures (\emph{where}) and requested operations (\emph{how}). In general-purpose code, an API-misuse diagnosis could select call expressions and their arguments for replacement. In configuration files, an incorrect-value diagnosis could select fields for replacement, whereas a missing-key diagnosis would request insertion without masking. Each domain would need identifiable constructs, its own grounding rules and correction order, and candidate checks such as tests, type checks, or schema validation. These extensions remain untested; Appendix~\ref{app:discussion} discusses their requirements.

\noindent\textbf{Limitations.}
We evaluate on one SQL benchmark, also used to select the correction order; predicted-type evaluation covers only single-error queries. TEG requires diagnoses and database access and adds inference cost. Execution accuracy does not guarantee semantic equivalence.

\subsection*{AI use statement}

AI tools assisted manuscript drafting and editing, consistency checks, and the development of analysis and experiment scripts, including ablations, aggregation, and diagnostics. They also assisted interpretation of experimental outputs. The authors reviewed the assisted work and take responsibility for the manuscript, code, and reported results.

\subsection*{Reproducibility statement}

The implementation of TEG, including the masking rules, the prompt templates, the error-type prediction prompt, the ablation scripts, and the evaluation script, together with the evaluation dataset, the BIRD development schemas, and the predicted error types, is available at \url{https://github.com/isuchan0212/why-where-how}.
The appendix documents the masking rules, prompts, evaluation dataset, baseline configurations, and hardware. Locally hosted models use greedy decoding, and execution accuracy is measured by set agreement of row tuples with the reference result.

\bibliographystyle{iclr2027_conference}
\bibliography{references}

\clearpage
\appendix
\etocdepthtag.toc{mtappendix}
\hypersetup{linkcolor=black}
\etocsettagdepth{mtchapter}{none}
\etocsettagdepth{mtappendix}{subsection}
\tableofcontents
\hypersetup{linkcolor=linkred}
\clearpage
\section{A Closer Look at TEG}
\label{app:implementation}

This section expands the description of TEG in Section~\ref{sec:method}. Sections~\ref{app:taxonomy} to~\ref{app:prompt-details} give the grounding rules behind the \emph{why}, the \emph{where}, and the \emph{how}: the taxonomy and its edit groups, the masking rules and algorithms for each error type, and the prompt templates that state each edit. Section~\ref{app:candidate-selection} covers candidate selection, which turns a grounded prompt into the corrected query by issuing it under the question and its paraphrases, executing the candidates, and keeping or refining one.

\subsection{Why: Error taxonomy and edit groups}
\label{app:taxonomy}

Table~\ref{tab:taxonomy} lists the 31 sub-types supported by the masking and prompt rules across nine categories. Of these, 29 occur in our 549-query NL2SQL-BUGs evaluation subset~\citep{liu2025nl2sql}. \textit{Join Type Mismatch} and \textit{Window Functions} have no annotations in this subset.
For prompt construction and the correction order, the implementation groups sub-types under \textit{Mismatch}~(M), \textit{Redundancy}~(R), \textit{Missing}~(Mi), or \textit{Rewrite}~(Rw). These groups summarize the requested edits. For sub-types such as the function errors, \textit{ASC/DESC}, and \textit{DISTINCT}, the prompt also depends on whether the corresponding construct is present and can be masked.
\textit{Mismatch} is the largest group, accounting for 64.7\% of all error annotations, followed by \textit{Missing} (21.3\%), \textit{Redundancy} (13.5\%), and \textit{Rewrite} (0.5\%).

\begin{table}[h]
\centering
\caption{\textbf{Four edit groups turn the 31 taxonomy sub-types into edit operations.}
    Sub-types follow the nine NL2SQL-BUGs categories.
    Edit groups: M = Mismatch (replace), R = Redundancy (remove), Mi = Missing (insert), Rw = Rewrite (restructure).
    Mismatch covers 64.7\% of all annotations, so replacement is the most frequent requested edit.}
\label{tab:taxonomy}
\small
\begin{tabular}{llc}
\toprule
\textbf{Category} & \textbf{Sub-type} & \textbf{Edit group} \\
\midrule
\multirow{3}{*}{Attribute-Related}
  & Attribute Mismatch     & M  \\
  & Attribute Redundancy   & R  \\
  & Attribute Missing      & Mi \\
\midrule
\multirow{5}{*}{Table-Related}
  & Table Mismatch         & M  \\
  & Table Redundancy       & R  \\
  & Table Missing          & Mi \\
  & Join Condition Mismatch & M \\
  & Join Type Mismatch      & M \\
\midrule
\multirow{2}{*}{Value-Related}
  & Value Mismatch         & M  \\
  & Data Format Mismatch   & M  \\
\midrule
\multirow{2}{*}{Operator-Related}
  & Comparison Operator Mismatch & M \\
  & Logical Operator Mismatch    & M \\
\midrule
\multirow{4}{*}{Condition-Related}
  & Explicit Condition Missing   & Mi \\
  & Implicit Condition Missing   & Mi \\
  & Explicit Condition Mismatch  & M  \\
  & Explicit Condition Redundancy & R \\
\midrule
\multirow{7}{*}{Function-Related}
  & Aggregate Functions    & M  \\
  & Window Functions       & M  \\
  & Date/Time Functions    & M  \\
  & Conversion Functions   & M  \\
  & Math Functions         & M  \\
  & String Functions       & M  \\
  & Conditional Functions  & M  \\
\midrule
\multirow{2}{*}{Clause-Related}
  & Clause Missing         & Mi \\
  & Clause Redundancy      & R  \\
\midrule
\multirow{3}{*}{Subquery-Related}
  & Subquery Missing       & Mi \\
  & Subquery Mismatch      & M  \\
  & Partial Query          & Rw \\
\midrule
\multirow{3}{*}{Other}
  & ASC/DESC               & M  \\
  & DISTINCT               & M  \\
  & Other                  & Rw \\
\bottomrule
\end{tabular}
\end{table}

\subsection{Where: Masking rules and algorithms}
\label{app:masking-rules}

Deciding which SQL constructs each error type implicates is part of the method: the taxonomy names the kind of error, and the rules below name the SQL constructs that can carry it. Table~\ref{tab:masking} gives that mapping for every error category and sub-type.
For maskable sub-types, SQLGlot traversal identifies matching node classes and replaces their occurrences with \texttt{<MASK>}. The annotation supplies an error type, not the identity of a unique faulty node.
\textit{Attribute-Related} masking extracts column expressions through \texttt{Column.sql()} and replaces their SQL strings, whereas the other categories use AST node replacement.
\textit{Table-Related} rows describe multi-error mode, the rules for queries with several annotations. Single-error mode, for queries with one annotation, masks both \texttt{Table} and \texttt{Join} nodes for every non-\textit{Missing} \textit{Table-Related} sub-type. Comparison-operator masks are restricted to \texttt{Where}/\texttt{Having}, and logical masks apply to outermost logical expressions. Generic function masks exclude the specialized function classes listed separately.
Entries marked with ``--'' correspond to sub-types in the \textit{Missing} or \textit{Rewrite} edit groups, which do not select existing constructs for masking. The query remains unmasked, and the edit instruction requests insertion or broader restructuring.

\begin{table}[h]
\centering
\caption{\textbf{Each maskable error sub-type selects the AST node classes that TEG masks.}
    Targets are SQLGlot node classes, and Table-Related rows describe multi-error mode.
    ``--'' marks sub-types in the Missing and Rewrite edit groups, which pass the query through unmasked.
    The rule names a node class rather than a unique faulty node, so the LLM still decides which masked construct to change.}
\label{tab:masking}
\small
\begin{tabular}{llp{6.2cm}}
\toprule
\textbf{Category} & \textbf{Sub-type} & \textbf{Target AST Nodes} \\
\midrule
\multirow{2}{*}{Attribute}
  & Mismatch, Redundancy & \texttt{Column} (string-level replace) \\
  & Missing              & -- \\
\midrule
\multirow{3}{*}{Table}
  & Mismatch, Redundancy         & \texttt{Table} \\
  & Join Cond./Type Mismatch     & \texttt{Join} \\
  & Missing                      & -- \\
\midrule
Value & Mismatch, Data Format & \texttt{Literal}, \texttt{String}, \texttt{Int64} \\
\midrule
\multirow{2}{*}{Operator}
  & Comparison Mismatch  & \texttt{EQ}, \texttt{GT}, \texttt{LT}, \texttt{GTE}, \texttt{LTE}, \texttt{NEQ}, \texttt{Like}, \texttt{ILike}, \texttt{In}, \texttt{Between}, \texttt{Is} \\
  & Logical Mismatch     & \texttt{And}, \texttt{Or}, \texttt{Not}, \texttt{Xor} \\
\midrule
\multirow{2}{*}{Condition}
  & Mismatch, Redundancy  & \texttt{Where}, plus \texttt{Having} in multi-error mode \\
  & Missing               & -- \\
\midrule
\multirow{5}{*}{Function}
  & Aggregate             & \texttt{AggFunc}, \texttt{Group}, \texttt{Having} \\
  & Window                & \texttt{Window} \\
  & Conversion            & \texttt{Cast} \\
  & Conditional           & \texttt{Case}, \texttt{If} \\
  & Date/Time, Math, String & \texttt{Func} \\
\midrule
\multirow{2}{*}{Clause}
  & Redundancy            & \texttt{Group}, \texttt{Order}, \texttt{Having}, \texttt{Limit} \\
  & Missing               & -- \\
\midrule
\multirow{2}{*}{Subquery}
  & Mismatch              & \texttt{Subquery} \\
  & Missing, Partial      & -- \\
\midrule
\multirow{3}{*}{Other}
  & ASC/DESC              & \texttt{Ordered} \\
  & DISTINCT              & \texttt{Distinct} \\
  & Other (Rw)            & -- \\
\bottomrule
\end{tabular}
\end{table}

\paragraph{Masking algorithms.}
\label{app:masking-procedures}

Algorithm~\ref{alg:masking} gives the generic procedure and Algorithms~\ref{alg:mask-clause} to~\ref{alg:mask-func} the per-category rules that decide which SQL constructs an error type implicates. Given an issued SQL $q$ and its error category and sub-type, TEG parses $q$ into an AST via SQLGlot, looks up the target node classes of Table~\ref{tab:masking}, and replaces every matching node with \texttt{<MASK>}.
For sub-types in the \textit{Missing} group and for \textit{Partial Query}, TEG returns the input query without parsing or masking. For \textit{Other}, it parses and serializes the query but selects no construct to mask. If this produces no placeholder and a previous correction step exists, TEG uses the retry template described in Section~\ref{app:prompt-details}.
All serialization uses the SQLite dialect, and parsing or masking exceptions return the query unchanged. Attribute masking operates on the SQL string as written in single-error mode and on the SQLGlot-serialized SQL in multi-error mode. Condition masking targets \texttt{Where} nodes and, in multi-error mode, \texttt{Having} nodes as well. Table masking distinguishes \texttt{Table} from \texttt{Join} targets in multi-error mode, whereas single-error mode masks both for every non-\textit{Missing} \textit{Table-Related} sub-type.
Two rules go beyond a plain node-class lookup. For \textit{Aggregate Functions}, TEG first masks \texttt{AggFunc}, \texttt{Group}, and \texttt{Having} nodes, then locates the first \texttt{Where} subtree in SQLGlot traversal order. Within that subtree, each comparison containing a placeholder is masked in full, so the LLM can regenerate the predicate together with the aggregate. Comparison masks are limited to \texttt{Where} and \texttt{Having}, and logical masks cover only the outermost logical expression, so nested operators are masked once.

\begin{algorithm}[h]
\caption{$\textsc{Masking}(q,\, t,\, s)$}
\label{alg:masking}
\begin{algorithmic}[1]
\Require SQL query $q$, error category $t$, sub-type $s$
\Ensure Masked query $\tilde{q}$
\If{$s$ is a \textit{Missing} sub-type or $s =$ \textit{Partial Query}}
    \State \Return $q$
\EndIf
\State $\mathrm{AST} \leftarrow \textsc{Parse}(q)$
\If{parse fails} \Return $q$ \EndIf
\State $\mathcal{N} \leftarrow \textsc{TargetNodes}(t,\, s)$
\If{$\mathcal{N} = \emptyset$}
    \State \Return $\mathrm{AST}.\textsc{ToSQL}()$ \Comment{Other: no mask}
\EndIf
\ForAll{node $v \in \mathrm{AST}$ matching $\mathcal{N}$}
    \State replace $v$ with \texttt{<MASK>}
\EndFor
\State \Return $\mathrm{AST}.\textsc{ToSQL}()$
\end{algorithmic}
\end{algorithm}

\begin{algorithm}[h]
\caption{$\textsc{MaskClause}(q,\, s)$}
\label{alg:mask-clause}
\begin{algorithmic}[1]
\Require SQL $q$, sub-type $s$
\If{$s =$ \textit{Clause Missing}}
    \State \Return $q$
\EndIf
\State $\mathrm{AST} \leftarrow \textsc{Parse}(q)$
\ForAll{$v \in \mathrm{AST}.\textsc{FindAll}(\texttt{Group},\, \texttt{Order},\, \texttt{Having},\, \texttt{Limit})$}
    \State $v.\textsc{Replace}(\texttt{<MASK>})$
\EndFor
\State \Return $\mathrm{AST}.\textsc{ToSQL}()$
\end{algorithmic}
\end{algorithm}

\begin{algorithm}[h]
\caption{$\textsc{MaskAttribute}(q,\, s)$}
\label{alg:mask-attr}
\begin{algorithmic}[1]
\Require SQL $q$, sub-type $s$
\If{$s =$ \textit{Attribute Missing}}
    \State \Return $q$
\EndIf
\State $\mathrm{AST} \leftarrow \textsc{Parse}(q)$
\State $\mathcal{C} \leftarrow \{c.\textsc{sql}() \mid c \in \mathrm{AST}.\textsc{FindAll}(\texttt{Column})\}$
\ForAll{$c \in \mathcal{C}$}
    \State $q \leftarrow q.\textsc{Replace}(c,\; \texttt{<MASK>})$ \Comment{String-level replacement}
\EndFor
\State \Return $q$
\end{algorithmic}
\end{algorithm}

\begin{algorithm}[h]
\caption{$\textsc{MaskSubquery}(q,\, s)$}
\label{alg:mask-subquery}
\begin{algorithmic}[1]
\Require SQL $q$, sub-type $s$
\If{$s \in \{\textit{Subquery Missing},\, \textit{Partial Query}\}$}
    \State \Return $q$
\EndIf
\State $\mathrm{AST} \leftarrow \textsc{Parse}(q)$
\ForAll{$v \in \mathrm{AST}.\textsc{FindAll}(\texttt{Subquery})$}
    \State $v.\textsc{Replace}(\texttt{<MASK>})$
\EndFor
\State \Return $\mathrm{AST}.\textsc{ToSQL}()$
\end{algorithmic}
\end{algorithm}

\begin{algorithm}[h]
\caption{$\textsc{MaskTable}(q,\, s)$}
\label{alg:mask-table}
\begin{algorithmic}[1]
\Require SQL $q$, sub-type $s$
\If{$s =$ \textit{Table Missing}}
    \State \Return $q$
\EndIf
\State $\mathrm{AST} \leftarrow \textsc{Parse}(q)$
\If{$s \in \{\textit{Table Mismatch},\, \textit{Table Redundancy}\}$} \Comment{Single-error mode masks both}
    \ForAll{$v \in \mathrm{AST}.\textsc{FindAll}(\texttt{Table})$}
        \State $v.\textsc{Replace}(\texttt{<MASK>})$
    \EndFor
\ElsIf{$s \in \{\textit{Join Condition Mismatch},\, \textit{Join Type Mismatch}\}$}
    \ForAll{$j \in \mathrm{AST}.\textsc{FindAll}(\texttt{Join})$}
        \State $j.\textsc{Replace}(\texttt{<MASK>})$
    \EndFor
\EndIf
\State \Return $\mathrm{AST}.\textsc{ToSQL}()$
\end{algorithmic}
\end{algorithm}

\begin{algorithm}[h]
\caption{$\textsc{MaskValue}(q,\, s)$}
\label{alg:mask-value}
\begin{algorithmic}[1]
\Require SQL $q$, sub-type $s$
\State $\mathrm{AST} \leftarrow \textsc{Parse}(q)$
\State $\mathcal{L} \leftarrow \mathrm{AST}.\textsc{FindAll}(\texttt{Literal},\, \texttt{String},\, \texttt{Int64})$
\ForAll{$\ell \in \mathcal{L}$}
    \State $\ell.\textsc{Replace}(\texttt{<MASK>})$
\EndFor
\State \Return $\mathrm{AST}.\textsc{ToSQL}()$
\end{algorithmic}
\end{algorithm}

\begin{algorithm}[h]
\caption{$\textsc{MaskOther}(q,\, s)$}
\label{alg:mask-other}
\begin{algorithmic}[1]
\Require SQL $q$, sub-type $s$
\State $\mathrm{AST} \leftarrow \textsc{Parse}(q)$
\If{$s =$ \textit{ASC/DESC}}
    \State Mask all \texttt{Ordered} nodes
\ElsIf{$s =$ \textit{DISTINCT}}
    \State Mask all \texttt{Distinct} nodes
\Else
    \State Leave the AST unmasked \Comment{Other (Rewrite)}
\EndIf
\State \Return $\mathrm{AST}.\textsc{ToSQL}()$
\end{algorithmic}
\end{algorithm}

\begin{algorithm}[h]
\caption{$\textsc{MaskOperator}(q,\, s)$}
\label{alg:mask-op}
\begin{algorithmic}[1]
\Require SQL $q$, sub-type $s$
\State $\mathrm{AST} \leftarrow \textsc{Parse}(q)$
\If{$s =$ \textit{Comparison Operator Mismatch}}
    \State $\mathcal{S} \leftarrow \mathrm{AST}.\textsc{FindAll}(\texttt{Where},\, \texttt{Having})$
    \ForAll{scope $\in \mathcal{S}$}
        \ForAll{comparison $c \in$ scope}
            \State $c.\textsc{Replace}(\texttt{<MASK>})$
        \EndFor
    \EndFor
\ElsIf{$s =$ \textit{Logical Operator Mismatch}}
    \ForAll{$v \in \mathrm{AST}.\textsc{FindAll}(\texttt{And},\, \texttt{Or},\, \texttt{Not},\, \texttt{Xor})$}
        \If{parent of $v$ is not a logical operator} \Comment{Avoid double-masking}
            \State $v.\textsc{Replace}(\texttt{<MASK>})$
        \EndIf
    \EndFor
\EndIf
\State \Return $\mathrm{AST}.\textsc{ToSQL}()$
\end{algorithmic}
\end{algorithm}

\begin{algorithm}[h]
\caption{$\textsc{MaskCondition}(q,\, s)$}
\label{alg:mask-cond}
\begin{algorithmic}[1]
\Require SQL $q$, sub-type $s$
\If{$s \in \{\textit{Explicit Condition Missing},\, \textit{Implicit Condition Missing}\}$}
    \State \Return $q$
\EndIf
\State $\mathrm{AST} \leftarrow \textsc{Parse}(q)$
\State $\mathcal{T} \leftarrow \{\texttt{Where}\}$, plus \texttt{Having} in multi-error mode
\ForAll{$w \in \mathrm{AST}.\textsc{FindAll}(\mathcal{T})$}
    \State $w.\textsc{Replace}(\texttt{<MASK>})$
\EndFor
\State \Return $\mathrm{AST}.\textsc{ToSQL}()$
\end{algorithmic}
\end{algorithm}

\begin{algorithm}[h]
\caption{$\textsc{MaskFunction}(q,\, s)$}
\label{alg:mask-func}
\begin{algorithmic}[1]
\Require SQL $q$, sub-type $s$
\State $\mathrm{AST} \leftarrow \textsc{Parse}(q)$
\If{$s =$ \textit{Aggregate Functions}}
    \State Mask all \texttt{AggFunc}, \texttt{Group}, \texttt{Having} nodes
    \State $w \leftarrow$ first \texttt{Where} subtree, if present
    \State Mask comparisons within $w$ containing \texttt{<MASK>}, if $w$ exists
\ElsIf{$s =$ \textit{Window Functions}}
    \State Mask all \texttt{Window} nodes
\ElsIf{$s =$ \textit{Conversion Functions}}
    \State Mask all \texttt{Cast} nodes
\ElsIf{$s =$ \textit{Conditional Functions}}
    \State Mask all \texttt{Case}, \texttt{If} nodes
\Else \Comment{Date/Time, Math, String Functions}
    \State Mask all \texttt{Func} nodes except \texttt{AggFunc}, \texttt{Cast}, \texttt{Convert}, \texttt{Case}, \texttt{If}, \texttt{Window}
\EndIf
\State \Return $\mathrm{AST}.\textsc{ToSQL}()$
\end{algorithmic}
\end{algorithm}

\FloatBarrier
\subsection{How: Prompt templates for each edit}
\label{app:prompt-details}

The correction prompt is constructed from four components: the (possibly masked) SQL query $\tilde{q}$, the natural language question $n$, the database schema $S$, and the error category and sub-type $(t, s)$.
TEG uses five correction templates, each covering one or more error categories, plus a retry template for multi-error steps whose mask is empty and a base template for queries without an annotation. The masked query replaces the issued query in the prompt, and no separate copy of the unmasked query is supplied. Every template that uses an annotation supplies its category and sub-type labels without the taxonomy definitions, and the model returns the corrected query as JSON: \texttt{\{"sql": "..."\}}.

\paragraph{Default template.}
Used for most categories (\textit{Attribute-Related}, \textit{Table-Related}, \textit{Value-Related}, \textit{Condition-Related}, \textit{Clause-Related}).
The instruction reflects the requested edit. For \textit{Mismatch} errors, the model is asked to fix the masked part. For \textit{Redundancy}, it is asked to delete the masked part. For \textit{Missing} errors, it is asked to insert the absent SQL fragment at the appropriate position.

\paragraph{Operator-Related template.}
If masked parts are present, the model fixes or deletes them. If no mask is present, the model inserts the appropriate operator (e.g., \texttt{AND}, \texttt{OR}, comparison operators).

\paragraph{Function-Related template.}
If masked parts are present, the model fixes or deletes them. If no mask is present, the model inserts the appropriate function (e.g., \texttt{CAST}, \texttt{CASE}, aggregate functions).

\paragraph{Subquery-Related template.}
The instruction follows the same \textit{Mismatch}/\textit{Redundancy}/\textit{Missing} structure as the default template, with two additions:
\textit{Missing} errors specifically prompt insertion of the missing subquery, and
\textit{Partial Query} errors prompt completion of an incomplete SQL query.

\paragraph{Other template.}
If masked parts are present, the model fixes or deletes them. If no mask is present, the behavior depends on the sub-type. \textit{ASC/DESC} errors prompt insertion of the missing fragment. \textit{DISTINCT} errors prompt a rewrite that may involve adding or removing the \texttt{DISTINCT} keyword, or restructuring joins, \texttt{SELECT} columns, or grouping. \textit{Other} errors prompt a rewrite from scratch unless the retry condition below applies.

\paragraph{Retry template.}
When a masking attempt yields no \texttt{<MASK>} and a previous correction step exists, the step uses this template with the unmasked current query. It lists the previous and current annotations, describes the first as a previously failed correction attempt and the second as the error to fix, and asks for a regenerated query. It carries no operation-specific instruction. At the first step, an empty mask falls back to the correction template for the current error category instead.

\paragraph{Base template.}
The base template supplies the query, question, and schema with no error information and asks the model to regenerate the query if it is semantically misaligned. It is the prompt of the Base rows in the tables, TEG's one-call fallback when no error type is predicted, and the correction prompt of the w/o error annotation ablation, which keeps paraphrases and execution-based selection.

\subsection{Execution-based candidate selection}
\label{app:candidate-selection}

Grounding determines the input of a correction step, and candidate selection decides its output.
A single answer to a grounded prompt can name a column or value that the database does not contain, so TEG generates several candidate queries, filters them by executing them, and asks an LLM to select or refine one, a stage developed from SQLFixAgent (Section~\ref{sec:step}).
The crafter is called once per query and asked for four paraphrases of the natural language question, of which the first three parsed paraphrases are kept, so each correction step issues its prompt at most four times, under the original question and under each paraphrase, and yields at most four candidate queries; every step reuses the same paraphrases with its current query and edit instruction.
Candidates are executed against the database with a 5s timeout per candidate. The execution filter prefers candidates that run, return at least one row, and contain no NULL in the first five returned rows. The filter is a heuristic, since valid results can be empty or contain NULL. If none passes, all generated candidates remain eligible, and duplicate SQL strings are then collapsed. With no generated candidate, TEG retains the current query. With one distinct candidate, it returns that candidate directly, even if it failed the execution filter.
With multiple candidates, the refiner receives the question, schema, candidate SQL strings, and accumulated failure messages, each recording the selected SQL, the SQLite error, and the exception class. The database-content field of the SQLFixAgent prompt is left empty. The refiner returns a reasoning string and one SQL, selected or rewritten, for up to three rounds, and stops when its SQL passes the same execution filter or times out. The timeout stop is inherited from the SQLFixAgent implementation and does not certify correctness. A round that returns no SQL is skipped. If no round meets either stopping condition, TEG returns the last candidate in the pool. Evaluation still scores timeouts as incorrect.
The crafter and refiner prompts are those of SQLFixAgent~\citep{cen2025sqlfixagent}, with its similar-example memory left uninitialized, which is the authors' no-memory branch.

\clearpage
\section{Experimental Details}
\label{app:setup}

This section extends Section~\ref{sec:setup} with the data, the Qwen3-8B thinking modes, the baseline implementations, the ablation variants of Section~\ref{sec:analysis}, and the hardware.

\subsection{Dataset statistics and examples}
\label{app:data-stats}

Our evaluation subset consists of 549 issued SQL queries from NL2SQL-BUGs~\citep{liu2025nl2sql}, derived from BIRD~\citep{li2023can} across 11 SQLite databases. We retain queries labeled incorrect that have a reference SQL.
Across this filtered subset, 1{,}160 error annotations cover nine categories and 29 observed sub-types. These are the sub-types observed after filtering, not the full dataset taxonomy.
Table~\ref{tab:data-stats} summarizes the distribution of errors by errors per query, category, and edit group.
Most queries (54.1\%) contain exactly two errors, and single-error queries are 20.0\%.

\begin{table}[h]
\centering\small
\caption{\textbf{Most evaluation queries carry more than one error.}
    The 1{,}160 error annotations in our 549-query NL2SQL-BUGs subset by errors per query, error category, and edit group.
    Single-error queries are 20.0\% of the subset, and Attribute and Table errors are about half of all annotations.}
\label{tab:data-stats}
\setlength{\tabcolsep}{6pt}
\begin{tabular}{@{}l r r@{}}
\toprule
\textbf{Distribution} & \textbf{Count} & \textbf{\%} \\
\midrule
\multicolumn{3}{@{}l}{\textsc{Errors per Query} \textit{(549 queries)}} \\[2pt]
\quad 1 error  & 110 & 20.0 \\
\quad 2 errors & 297 & 54.1 \\
\quad 3 errors & 114 & 20.8 \\
\quad 4 errors &  26 &  4.7 \\
\quad 5 errors &   2 &  0.4 \\
\midrule
\multicolumn{3}{@{}l}{\textsc{Error Category} \textit{(1{,}160 annotations)}} \\[2pt]
\quad Attribute  & 340 & 29.3 \\
\quad Table      & 247 & 21.3 \\
\quad Condition  & 166 & 14.3 \\
\quad Value      & 132 & 11.4 \\
\quad Function   &  88 &  7.6 \\
\quad Clause     &  83 &  7.2 \\
\quad Other      &  58 &  5.0 \\
\quad Subquery   &  26 &  2.2 \\
\quad Operator   &  20 &  1.7 \\
\midrule
\multicolumn{3}{@{}l}{\textsc{Edit Group} \textit{(1{,}160 annotations)}} \\[2pt]
\quad Mismatch (M)    & 750 & 64.7 \\
\quad Redundancy (R)  & 157 & 13.5 \\
\quad Missing (Mi)    & 247 & 21.3 \\
\quad Rewrite (Rw)    &   6 &  0.5 \\
\bottomrule
\end{tabular}
\end{table}

\paragraph{Examples.}
Table~\ref{tab:data-examples} shows representative single-error and multi-error cases from several error categories, with the erroneous and reference spans marked.

\begin{table}[h]
\centering\small
\caption{\textbf{An error can sit in a column, a value, or several places at once.}
    Representative single-error and multi-error queries from NL2SQL-BUGs.
    Red marks the erroneous span in the issued SQL and green the corresponding span in the reference SQL.
    Example 3 needs both an insertion and a replacement, which TEG handles in two sequential steps.}
\label{tab:data-examples}
\begin{tabularx}{\textwidth}{p{2.2cm}X}
\toprule
\multicolumn{2}{l}{\textbf{Example 1}: Attribute Mismatch (single error) \hfill \texttt{db: california\_schools}} \\
\midrule
Question & What is the unabbreviated mailing street address of the school with the highest FRPM count for K-12 students? \\
Issued SQL & \texttt{SELECT schools.\colorbox{red!15}{streetabr} FROM frpm INNER JOIN schools ON frpm.cdscode = schools.cdscode ORDER BY frpm.`frpm count (k-12)` DESC LIMIT 1} \\
Reference SQL & \texttt{SELECT T2.\colorbox{green!15}{MailStreet} FROM frpm AS T1 INNER JOIN schools AS T2 ON T1.CDSCode = T2.CDSCode ORDER BY T1.`FRPM Count (K-12)` DESC LIMIT 1} \\
\midrule\midrule
\multicolumn{2}{l}{\textbf{Example 2}: Value Mismatch (single error) \hfill \texttt{db: codebase\_community}} \\
\midrule
Question & Which user has the website URL listed at 'http://stackoverflow.com' \\
Issued SQL & \texttt{SELECT displayname FROM users WHERE websiteurl = \colorbox{red!15}{'http://stackoverflow.com/u/1114'}} \\
Reference SQL & \texttt{SELECT displayname FROM users WHERE websiteurl = \colorbox{green!15}{'http://stackoverflow.com'}} \\
\midrule\midrule
\multicolumn{2}{l}{\textbf{Example 3}: Attribute Missing + Attribute Mismatch (multi-error) \hfill \texttt{db: california\_schools}} \\
\midrule
Question & Please specify all of the schools and their related mailing zip codes that are under Avetik Atoian's administration. \\
Issued SQL & \texttt{SELECT mailzip FROM schools WHERE admfname1 = 'Avetik' AND \colorbox{red!15}{admlname2} = 'Atoian'} \\
Reference SQL & \texttt{SELECT \colorbox{green!15}{school,} mailzip FROM schools WHERE \colorbox{green!15}{admlname1} = 'Atoian' AND admfname1 = 'Avetik'} \\
\bottomrule
\end{tabularx}
\end{table}

\subsection{Qwen3-8B thinking modes}
\label{app:qwen3-thinking}

Qwen3-8B runs with thinking OFF and ON in every table. Both modes use greedy decoding and the same correction procedures. OFF disables thinking in the chat template and permits up to 4{,}096 generated tokens per model call. ON adds a 4{,}096-token thinking budget, for a total generation limit of 8{,}192 tokens. The implementation forces the end-of-thinking token when that budget is reached. The modes therefore differ in their available generation budget.

Every method improves with thinking ON, in the main comparison (Table~\ref{tab:baseline_single}), with predicted types (Table~\ref{tab:predicted_single}), and with baselines given error types (Table~\ref{tab:baseline_annotated}), and TEG leads in both modes in each.

\subsection{Baseline implementations}
\label{app:baselines}

The baselines share TEG's LLM and decoding settings within each comparison column. Base and the reasoning and SQL-specific baselines are evaluated both with and without oracle error types. ErrorLLM always receives the same annotations as TEG.
We re-implemented the methods and chose iteration counts in preliminary experiments.
TEG's crafter and refiner reuse the SQLFixAgent authors' prompts, whereas the reported SQLFixAgent baseline uses our revised prompts. The reported rows are as follows.

\begin{itemize}
    \item \textbf{CoT}~\citep{kojima2022large}: The model is prompted to analyze what the question requires, where the given SQL deviates from it, and what fix is needed, then return its reasoning and corrected SQL. The main comparison uses no error-type information or masking.

    \item \textbf{DIN-SQL}~\citep{pourreza2023din}: A full NL2SQL pipeline that decomposes the task into schema linking, query classification, and SQL generation, using the issued SQL as context. The generated SQL is then reviewed and revised via a self-correction stage. The schema context includes up to three sample rows per table when available; generated queries are not executed for correction feedback.

    \item \textbf{Self-Debug (Simple)}~\citep{chen2024teaching}: The simple-feedback variant, which uses no execution signal. After an initial single-pass correction, each turn asks the model to judge whether its current query is correct and, if not, to emit a revision. We allow up to 3 feedback-correction turns and stop early as soon as the model judges the current query correct.

    \item \textbf{Self-Debug (Expl)}~\citep{chen2024teaching}: Adds execution feedback and self-explanation on top of the simple variant. Each turn executes the current query and issues three calls: one explaining what the query returns given its execution result, one inferring what the question asks for, and one comparing the two to judge and revise. Up to 3 turns with the same early stopping.

    \item \textbf{MAC-SQL}~\citep{wang2025mac}: Our adaptation supplies the issued SQL to a Decomposer that decomposes the question and generates corrected SQL. A Selector prunes the schema only when its representation exceeds 3,000 tokens, counted with a Qwen2.5 tokenizer. A Refiner revises the generated SQL on execution errors, timeouts, or empty results for up to 3 rounds, stopping when execution returns a nonempty result.

    \item \textbf{SQLFixAgent}~\citep{cen2025sqlfixagent}: A multi-agent pipeline that first repairs the issued SQL in a single pass, then prompts a reviewer to check the result clause by clause for semantic errors. Its error check is prompt-based and uses no execution signal, and queries the reviewer passes are returned unchanged. When an error is flagged, a crafter produces up to 3 question paraphrases, and a separate generation call for each paraphrase uses the issued SQL to produce a candidate, keeping those that execute without error, and a refiner then selects or rewrites among the candidates using execution-based verification with a failure memory, for up to 3 rounds. The candidate filter accepts any executable query, including empty results; the refiner stops on a nonempty execution result and records execution failures or empty results in its failure memory for the next attempt. TEG additionally prefers nonempty results without NULL in the first five rows, and keeps all candidates when none passes.

    \item \textbf{ErrorLLM}~\citep{hong2026errorllm}: We use a structural refinement adaptation with the same LLM and supplied annotations as TEG. A localization call receives the question, schema, SQLGlot AST with node IDs, and error-specific guidelines, and returns relevant AST nodes, schema elements, and filled analyses. The implementation extracts the corresponding AST subtrees and related schema context. A second call receives these contexts, the filled guidelines, correction examples, and issued SQL, and corrects the supplied errors jointly in priority order. The schema graph uses database metadata and lexical links to the question, rather than the original ELECTRA/RGAT QSS encoder. We do not reproduce the trained error-token detector. When no valid AST-node or schema-element references remain, context extraction falls back to the full AST or schema graph, respectively, and invalid refinement output or an empty prediction retains the issued query. The method uses no database execution feedback during generation.
\end{itemize}

\subsection{Ablation variants}
\label{app:ablation-variants}

Table~\ref{tab:analysis} removes or alters one part of TEG per row, with Qwen2.5-7B-Instruct and oracle error types. The variants are implemented as follows.
The w/o error annotation ablation drops the annotation and therefore masking and sequential correction as well: it issues one correction step with the base template of Section~\ref{app:prompt-details}, retaining paraphrases and execution-based candidate selection.
The w/o masking ablation keeps the annotations and category-specific templates but leaves the query unmasked and disables the retry branch. Removing the masks can also change which conditional edit instruction applies.
The w/o DB execution ablation removes database execution during correction. Candidates are de-duplicated without execution-based filtering. With multiple distinct candidates, the refiner is called once with an empty failure memory, falling back to the last candidate if it returns no SQL. A sole candidate is returned directly; if none is generated, the current query is retained.
The w/o candidate selection ablation removes the crafter and the refiner, leaving one prompt and one candidate per correction step.
The w/o sequential correction ablation applies the masks of all annotations cumulatively to one query and issues a single prompt built from the template of the first annotation's category, listing every annotation. The candidate selection stage is unchanged and runs once on that single prompt.

\subsection{Computing resources}
\label{app:computing}

The locally hosted Qwen2.5 experiments were conducted on a compute node with an AMD EPYC 9654 96-core processor and four NVIDIA RTX PRO 6000 Blackwell GPUs (96 GB VRAM each). Each local experimental run used a single dedicated GPU. Opus~4.8 was accessed through an API on hardware we did not control.

\clearpage
\section{Additional Main Results}
\label{app:main-comparison}

This section extends Section~\ref{sec:main} with per-category single-error and multi-error results and the full results for the baselines given oracle error types.

\subsection{Single-error results by error category}
\label{app:error-category-results}

Tables~\ref{tab:error-categories-qwen25} and~\ref{tab:error-categories-qwen3} break down the main single-error comparison by the oracle error category for all five Qwen settings. Access to error types matches the main comparison: Base, CoT, Self-Debug, DIN-SQL, MAC-SQL, and SQLFixAgent receive no error types, while ErrorLLM and TEG receive oracle error types. Each block uses the same single-error queries, comprising Attribute, Value, Condition, Table, and Others. Others combines \textit{Operator-Related}, \textit{Function-Related}, \textit{Clause-Related}, and catch-all Other errors. There are no \textit{Subquery-Related} queries in the single-error subset. Overall accuracy weights queries equally rather than averaging the category accuracies.

The relative performance varies by category. TEG leads or ties on Attribute errors in all five Qwen settings. DIN-SQL is stronger on Value errors at Qwen2.5-14B and 32B. ErrorLLM leads on Condition and Table errors at 14B with 70.6 and 73.3, compared with 58.8 and 66.7 for TEG, despite TEG's higher overall accuracy. The small Table and Others subsets contain 15 and 12 queries, respectively, so a difference of one or two queries can change their category rankings.

\begin{table}[h]
\centering\footnotesize
\caption{\textbf{\sname{} leads or ties on Attribute errors at every Qwen2.5 scale.}
    Single-error execution accuracy by oracle error category on Qwen2.5-Instruct.
    In the Error types column, \ding{51} denotes supplied oracle error types from NL2SQL-BUGs; \ding{55} denotes no error types.
    Others pools the less frequent categories, and Overall counts all queries.
    Bold marks the best value per column within each model block, including ties.
    Other methods win smaller categories, such as DIN-SQL on Value errors at 14B and 32B.}
\label{tab:error-categories-qwen25}
\setlength{\tabcolsep}{4pt}
\begin{tabular}{l|c|ccccc|c}
\toprule
Method & Error types & Attribute & Value & Condition & Table & Others & Overall \\
\midrule
\family{8}{Qwen2.5-7B} \\
\addlinespace[0.2em]
Base & \ding{55} & 24.3 & 41.4 & 29.4 & 33.3 & 16.7 & 30.0 \\
CoT & \ding{55} & 35.1 & 37.9 & 52.9 & 26.7 & 16.7 & 35.5 \\
Self-Debug (Simple) & \ding{55} & 24.3 & 41.4 & 35.3 & 53.3 & 16.7 & 33.6 \\
Self-Debug (Expl) & \ding{55} & 24.3 & 37.9 & 35.3 & 33.3 & 8.3 & 29.1 \\
DIN-SQL & \ding{55} & 10.8 & \textbf{51.7} & 23.5 & 13.3 & 16.7 & 24.5 \\
MAC-SQL & \ding{55} & 10.8 & 27.6 & 23.5 & 6.7 & \textbf{25.0} & 18.2 \\
SQLFixAgent & \ding{55} & 24.3 & 41.4 & 41.2 & 40.0 & 16.7 & 32.7 \\
ErrorLLM & \ding{51} & 16.2 & 44.8 & 41.2 & 60.0 & 16.7 & 33.6 \\
\rowcolor{tabtint} \sname (ours) & \ding{51} & \textbf{37.8} & \textbf{51.7} & \textbf{58.8} & \textbf{66.7} & \textbf{25.0} & \textbf{47.3} \\
\midrule
\family{8}{Qwen2.5-14B} \\
\addlinespace[0.2em]
Base & \ding{55} & 27.0 & 37.9 & 41.2 & 40.0 & \textbf{41.7} & 35.5 \\
CoT & \ding{55} & 24.3 & 37.9 & 35.3 & 66.7 & \textbf{41.7} & 37.3 \\
Self-Debug (Simple) & \ding{55} & 37.8 & 37.9 & 47.1 & 40.0 & 25.0 & 38.2 \\
Self-Debug (Expl) & \ding{55} & 29.7 & 41.4 & 29.4 & 40.0 & \textbf{41.7} & 35.5 \\
DIN-SQL & \ding{55} & 24.3 & \textbf{65.5} & 58.8 & 33.3 & 33.3 & 42.7 \\
MAC-SQL & \ding{55} & 8.1 & 31.0 & 11.8 & 0.0 & 25.0 & 15.5 \\
SQLFixAgent & \ding{55} & 29.7 & 41.4 & 35.3 & 46.7 & 33.3 & 36.4 \\
ErrorLLM & \ding{51} & 43.2 & 48.3 & \textbf{70.6} & \textbf{73.3} & 33.3 & 51.8 \\
\rowcolor{tabtint} \sname (ours) & \ding{51} & \textbf{48.6} & 55.2 & 58.8 & 66.7 & 33.3 & \textbf{52.7} \\
\midrule
\family{8}{Qwen2.5-32B} \\
\addlinespace[0.2em]
Base & \ding{55} & 32.4 & 48.3 & 35.3 & 66.7 & 33.3 & 41.8 \\
CoT & \ding{55} & 40.5 & 37.9 & 47.1 & 80.0 & 33.3 & 45.5 \\
Self-Debug (Simple) & \ding{55} & 40.5 & 55.2 & 41.2 & 66.7 & \textbf{41.7} & 48.2 \\
Self-Debug (Expl) & \ding{55} & 43.2 & 44.8 & 35.3 & 66.7 & 33.3 & 44.5 \\
DIN-SQL & \ding{55} & 35.1 & \textbf{79.3} & \textbf{58.8} & 53.3 & \textbf{41.7} & 53.6 \\
MAC-SQL & \ding{55} & 13.5 & 27.6 & 0.0 & 26.7 & 25.0 & 18.2 \\
SQLFixAgent & \ding{55} & 37.8 & 48.3 & 35.3 & 66.7 & 33.3 & 43.6 \\
ErrorLLM & \ding{51} & 43.2 & 51.7 & \textbf{58.8} & 86.7 & \textbf{41.7} & 53.6 \\
\rowcolor{tabtint} \sname (ours) & \ding{51} & \textbf{56.8} & 55.2 & 52.9 & \textbf{93.3} & 25.0 & \textbf{57.3} \\
\bottomrule
\end{tabular}
\end{table}

\clearpage
\begin{table}[h]
\centering\footnotesize
\caption{\textbf{On Qwen3-8B, \sname{} leads or ties on Attribute errors with thinking both OFF and ON.}
    Single-error execution accuracy by oracle error category.
    In the Error types column, \ding{51} denotes supplied oracle error types from NL2SQL-BUGs; \ding{55} denotes no error types.
    Others pools the less frequent categories, and Overall weights queries equally, not category columns.
    Bold marks the best value per column within each thinking mode, including ties.
    Other methods win smaller categories, such as Self-Debug (Simple) on Value errors with thinking ON.}
\label{tab:error-categories-qwen3}
\setlength{\tabcolsep}{4pt}
\begin{tabular}{l|c|ccccc|c}
\toprule
Method & Error types & Attribute & Value & Condition & Table & Others & Overall \\
\midrule
\family{8}{Qwen3-8B (Thinking OFF)} \\
\addlinespace[0.2em]
Base & \ding{55} & 27.0 & 34.5 & 29.4 & 46.7 & 25.0 & 31.8 \\
CoT & \ding{55} & 18.9 & 37.9 & 41.2 & 66.7 & 16.7 & 33.6 \\
Self-Debug (Simple) & \ding{55} & 40.5 & 44.8 & 29.4 & 33.3 & \textbf{33.3} & 38.2 \\
Self-Debug (Expl) & \ding{55} & \textbf{43.2} & 44.8 & 35.3 & 66.7 & \textbf{33.3} & 44.5 \\
DIN-SQL & \ding{55} & 13.5 & 41.4 & 17.6 & 13.3 & 16.7 & 21.8 \\
MAC-SQL & \ding{55} & 8.1 & 37.9 & \textbf{47.1} & 26.7 & 16.7 & 25.5 \\
SQLFixAgent & \ding{55} & 32.4 & 44.8 & 35.3 & 46.7 & 25.0 & 37.3 \\
ErrorLLM & \ding{51} & 24.3 & 44.8 & 41.2 & 66.7 & \textbf{33.3} & 39.1 \\
\rowcolor{tabtint} \sname (ours) & \ding{51} & \textbf{43.2} & \textbf{48.3} & 41.2 & \textbf{80.0} & 16.7 & \textbf{46.4} \\
\midrule
\family{8}{Qwen3-8B (Thinking ON)} \\
\addlinespace[0.2em]
Base & \ding{55} & 37.8 & 44.8 & 35.3 & 73.3 & 33.3 & 43.6 \\
CoT & \ding{55} & 45.9 & 44.8 & 41.2 & \textbf{80.0} & 33.3 & 48.2 \\
Self-Debug (Simple) & \ding{55} & 45.9 & \textbf{55.2} & 35.3 & \textbf{80.0} & \textbf{41.7} & 50.9 \\
Self-Debug (Expl) & \ding{55} & 48.6 & 44.8 & 41.2 & 60.0 & \textbf{41.7} & 47.3 \\
DIN-SQL & \ding{55} & 29.7 & \textbf{55.2} & 41.2 & 60.0 & \textbf{41.7} & 43.6 \\
MAC-SQL & \ding{55} & 18.9 & 41.4 & 47.1 & 26.7 & 16.7 & 30.0 \\
SQLFixAgent & \ding{55} & 32.4 & 41.4 & 41.2 & 73.3 & \textbf{41.7} & 42.7 \\
ErrorLLM & \ding{51} & 29.7 & 48.3 & 47.1 & \textbf{80.0} & \textbf{41.7} & 45.5 \\
\rowcolor{tabtint} \sname (ours) & \ding{51} & \textbf{62.2} & 48.3 & \textbf{58.8} & 73.3 & \textbf{41.7} & \textbf{57.3} \\
\bottomrule
\end{tabular}
\end{table}

\subsection{Multi-error results by error category}
\label{app:multi-by-category}

Table~\ref{tab:multi_by_category} reports the 439 multi-error queries by category. TEG leads every reported category and reaches 34.4 overall. A query can occur in several category columns, so the columns overlap.

\begin{table}[h]
\centering\small
\caption{\textbf{On multi-error queries, \sname{} leads every error category and reaches 34.4 overall.}
    Execution accuracy by error category on Qwen2.5-7B-Instruct over the 439 multi-error queries.
    ErrorLLM and \sname{} use oracle types, while the other baselines run without them.
    A query counts in every category containing one of its errors, so the columns overlap (Attribute 275, Value 103, Condition 140, Table 194, Others 213), and bold marks the best value per column.
    The lead is largest on Table errors, 43.8 against 28.4 for the next best method.}
\label{tab:multi_by_category}
\setlength{\tabcolsep}{5.5pt}
\begin{tabular}{l|ccccc|c}
\toprule
Method & Attribute & Value & Condition & Table & Others & Overall \\
\midrule
Base & 19.6 & 19.4 & 15.7 & 24.2 & 22.1 & 22.1 \\
\midrule \addlinespace[0.3em]
\family{7}{Reasoning and self-correction} \\
\addlinespace[0.3em]
CoT~\citep{kojima2022large} & 20.7 & 13.6 & 15.7 & 28.4 & 21.6 & 21.6 \\
Self-Debug (Simple)~\citep{chen2024teaching} & 16.7 & 19.4 & 17.1 & 23.2 & 19.7 & 19.6 \\
Self-Debug (Expl)~\citep{chen2024teaching} & 22.2 & 17.5 & 15.7 & 25.3 & 22.1 & 22.6 \\
\midrule \addlinespace[0.3em]
\family{7}{SQL-specific refinement} \\
\addlinespace[0.3em]
DIN-SQL~\citep{pourreza2023din} & 18.5 & 22.3 & 13.6 & 18.6 & 20.2 & 18.7 \\
MAC-SQL~\citep{wang2025mac} & 11.3 & 12.6 & 11.4 & 12.9 & 11.3 & 12.3 \\
SQLFixAgent~\citep{cen2025sqlfixagent} & 23.6 & 20.4 & 17.9 & 28.4 & 24.4 & 24.6 \\
\midrule \addlinespace[0.3em]
\family{7}{Taxonomy-guided refinement} \\
\addlinespace[0.3em]
ErrorLLM~\citep{hong2026errorllm} & 20.7 & 17.5 & 15.7 & 27.8 & 17.8 & 21.6 \\
\midrule \addlinespace[0.3em]
\family{7}{Taxonomy-guided error grounding} \\
\addlinespace[0.3em]
\sname (ours) & \textbf{31.3} & \textbf{28.2} & \textbf{27.9} & \textbf{43.8} & \textbf{31.0} & \textbf{34.4} \\
\bottomrule
\end{tabular}
\end{table}

\subsection{Baselines given oracle error types}
\label{app:annotated-baselines}

This subsection reports in full the comparisons summarized in Section~\ref{sec:main}. ErrorLLM and TEG use oracle error types, whereas the other baselines in the main comparison do not. Here every baseline receives the same oracle error types, and TEG stays the most accurate in every Qwen setting.

\subsubsection{Single-error results}

Table~\ref{tab:baseline_annotated} reports the single-error comparisons on the same 110 queries and five Qwen settings as Table~\ref{tab:baseline_single}, with every method receiving the oracle error types that ErrorLLM and TEG use.

We add the error types to the prompts that generate or revise the issued SQL, retaining each method's correction procedure, iteration limits, and decoding settings. Base and CoT receive the error types directly in their correction prompts. Self-Debug receives them in its initial correction and feedback prompts. For DIN-SQL and MAC-SQL, we add them to SQL generation and decomposition, respectively. SQLFixAgent receives them in its initial correction and candidate-generation prompts. These changes supply the error types without adding TEG's type-specific masking or sequential correction. ErrorLLM and TEG already use these error types, so their rows are unchanged.

\begin{table}[h]
\centering\small
\caption{\textbf{\sname{} leads single-error correction when all methods receive oracle types.}
    Single-error execution accuracy.
    Base, reasoning and self-correction, and SQL-specific refinement methods use prompts augmented with those types, and ErrorLLM and \sname{} already use the types in their own procedures.
    Qwen3-8B is evaluated with thinking OFF and ON, and bold marks the best accuracy per Qwen setting.
    The single-error margin over the strongest baseline given error types ranges from 0.9 points, at Qwen2.5-14B and at Qwen3-8B with thinking OFF, to 11.8 points at Qwen2.5-7B.}
\label{tab:baseline_annotated}
\setlength{\tabcolsep}{5pt}
\begin{tabular}{l|ccc|cc}
\toprule
& \multicolumn{3}{c|}{\textbf{Qwen2.5-Instruct}} & \multicolumn{2}{c}{\textbf{Qwen3-8B (thinking)}} \\
Method & 7B & 14B & 32B & OFF & ON \\
\midrule
Base & 28.2 & 36.4 & 41.8 & 35.5 & 48.2 \\
\midrule \addlinespace[0.3em]
\family{6}{Reasoning and self-correction} \\
\addlinespace[0.3em]
CoT~\citep{kojima2022large} & 32.7 & 40.9 & 42.7 & 39.1 & 52.7 \\
Self-Debug (Simple)~\citep{chen2024teaching} & 35.5 & 45.5 & 50.9 & 40.9 & 48.2 \\
Self-Debug (Expl)~\citep{chen2024teaching} & 28.2 & 38.2 & 45.5 & 45.5 & 48.2 \\
\midrule \addlinespace[0.3em]
\family{6}{SQL-specific refinement} \\
\addlinespace[0.3em]
DIN-SQL~\citep{pourreza2023din} & 26.4 & 40.9 & 52.7 & 24.5 & 46.4 \\
MAC-SQL~\citep{wang2025mac} & 17.3 & 17.3 & 26.4 & 11.8 & 24.5 \\
SQLFixAgent~\citep{cen2025sqlfixagent} & 32.7 & 41.8 & 46.4 & 41.8 & 48.2 \\
\midrule \addlinespace[0.3em]
\family{6}{Taxonomy-guided refinement} \\
\addlinespace[0.3em]
ErrorLLM~\citep{hong2026errorllm} & 33.6 & 51.8 & 53.6 & 39.1 & 45.5 \\
\midrule \addlinespace[0.3em]
\family{6}{Taxonomy-guided error grounding} \\
\addlinespace[0.3em]
\sname (ours) & \textbf{47.3} & \textbf{52.7} & \textbf{57.3} & \textbf{46.4} & \textbf{57.3} \\
\bottomrule
\end{tabular}
\end{table}

TEG remains above every baseline given error types in all five Qwen settings. On Qwen2.5, Self-Debug (Simple) is the strongest such baseline at 7B with 35.5, while ErrorLLM is strongest at 14B and 32B with 51.8 and 53.6. Giving CoT the same types yields 32.7, 40.9, and 42.7 at 7B, 14B, and 32B, compared with 47.3, 52.7, and 57.3 for TEG. On Qwen3-8B, the strongest such baselines are Self-Debug (Expl) with thinking OFF at 45.5 and CoT with thinking ON at 52.7, compared with TEG at 46.4 and 57.3. Access to error types alone therefore does not account for TEG's advantage. The procedures also differ in their candidate selection stages, including generation and refinement, so these comparisons should be read together with the component ablations when assessing error grounding.

\subsubsection{Results by error count}
\label{app:annotated-error-count}

We apply the same prompt augmentation to all 549 queries with Qwen2.5-7B-Instruct. Table~\ref{tab:error_count_annotated} reports execution accuracy by error count when every method receives the same oracle error types. The correction procedures, iteration limits, and decoding settings are retained. ErrorLLM and TEG use annotations in their standard procedures, so their results are repeated from the main comparison.

\begin{table}[h]
\centering\small
\caption{\textbf{With the same oracle error types for every baseline, \sname{} leads at every error count.}
    Execution accuracy on all 549 queries using Qwen2.5-7B-Instruct.
    Base, reasoning and self-correction, and SQL-specific refinement methods use prompts augmented with those types, and ErrorLLM and \sname{} already use the types in their own procedures.
    Bold marks the best value per column.
    \sname{} reaches 37.0 overall against 26.0 for the strongest such baseline, Self-Debug (Simple).}
\label{tab:error_count_annotated}
\setlength{\tabcolsep}{4pt}
\begin{tabular}{l|ccc|c}
\toprule
Method & 1 error & 2 errors & 3+ errors & Overall \\
\midrule
Base & 28.2 & 23.2 & 16.9 & 22.6 \\
\midrule \addlinespace[0.3em]
\family{5}{Reasoning and self-correction} \\
\addlinespace[0.3em]
CoT~\citep{kojima2022large} & 32.7 & 25.9 & 14.8 & 24.4 \\
Self-Debug (Simple)~\citep{chen2024teaching} & 35.5 & 24.6 & 21.8 & 26.0 \\
Self-Debug (Expl)~\citep{chen2024teaching} & 28.2 & 26.9 & 19.7 & 25.3 \\
\midrule \addlinespace[0.3em]
\family{5}{SQL-specific refinement} \\
\addlinespace[0.3em]
DIN-SQL~\citep{pourreza2023din} & 26.4 & 19.9 & 15.5 & 20.0 \\
MAC-SQL~\citep{wang2025mac} & 17.3 & 10.4 & \phantom{0}9.2 & 11.5 \\
SQLFixAgent~\citep{cen2025sqlfixagent} & 32.7 & 26.3 & 19.0 & 25.7 \\
\midrule \addlinespace[0.3em]
\family{5}{Taxonomy-guided refinement} \\
\addlinespace[0.3em]
ErrorLLM~\citep{hong2026errorllm} & 33.6 & 23.6 & 17.6 & 24.0 \\
\midrule \addlinespace[0.3em]
\family{5}{Taxonomy-guided error grounding} \\
\addlinespace[0.3em]
\sname (ours) & \textbf{47.3} & \textbf{36.7} & \textbf{29.6} & \textbf{37.0} \\
\bottomrule
\end{tabular}
\end{table}

TEG reaches 37.0 overall, compared with 24.0 for ErrorLLM and 26.0 for Self-Debug (Simple), the strongest method among the baselines augmented with error types. TEG remains above every such baseline in each error-count group. These comparisons show that its advantage persists when all methods receive the same error types, while their correction procedures still differ.

\subsubsection{Single-error results by error category}
\label{app:baseline-by-category}

Table~\ref{tab:baseline_by_category} compares the Qwen2.5-7B methods with and without oracle types by error category. The 110 single-error queries comprise Attribute (37), Value (29), Condition (17), Table (15), and Others (12). Others combines Operator-, Function-, \textit{Clause-Related}, and catch-all Other errors. No \textit{Subquery-Related} errors occur in this single-error subset. %
Among the six reasoning and SQL-specific baselines, oracle types improve two, reduce three, and leave one unchanged, and they lower Base from 30.0 to 28.2. DIN-SQL with error types leads on Value errors, while MAC-SQL and TEG tie on Others. TEG has the highest overall accuracy, 47.3 against 35.5 for the best baseline given error types.

\begin{table}[h]
\centering\footnotesize
\caption{\textbf{Oracle types lift only two of six baselines, and \sname{} stays highest overall.}
    Single-error execution accuracy by error category on Qwen2.5-7B-Instruct.
    Each pair shows results without and with the oracle error type.
    ErrorLLM and \sname{} require types, so their entries without types are dashes.
    Bold marks the best value per category with and without oracle error types, including ties.
    With error types, DIN-SQL leads on Value, and SQLFixAgent and MAC-SQL tie \sname{} on Table and Others.}
\label{tab:baseline_by_category}
\setlength{\tabcolsep}{2.5pt}
\begin{tabular}{l|cc|cc|cc|cc|cc|cc}
\toprule
Method & \multicolumn{2}{c|}{Attribute} & \multicolumn{2}{c|}{Value} & \multicolumn{2}{c|}{Condition} & \multicolumn{2}{c|}{Table} & \multicolumn{2}{c|}{Others} & \multicolumn{2}{c}{Overall} \\
\midrule
Base & 24.3 & 18.9 & 41.4 & 37.9 & 29.4 & 29.4 & 33.3 & 46.7 & 16.7 & \phantom{0}8.3 & 30.0 & 28.2 \\
\midrule \addlinespace[0.3em]
\family{13}{Reasoning and self-correction} \\
\addlinespace[0.3em]
CoT & 35.1 & 21.6 & 37.9 & 34.5 & 52.9 & 47.1 & 26.7 & 53.3 & 16.7 & 16.7 & 35.5 & 32.7 \\
Self-Debug (Simple) & 24.3 & 24.3 & 41.4 & 48.3 & 35.3 & 41.2 & 53.3 & 46.7 & 16.7 & 16.7 & 33.6 & 35.5 \\
Self-Debug (Expl) & 24.3 & 16.2 & 37.9 & 31.0 & 35.3 & 35.3 & 33.3 & 53.3 & \phantom{0}8.3 & 16.7 & 29.1 & 28.2 \\
\midrule \addlinespace[0.3em]
\family{13}{SQL-specific refinement} \\
\addlinespace[0.3em]
DIN-SQL & 10.8 & 10.8 & 51.7 & \textbf{55.2} & 23.5 & 29.4 & 13.3 & 13.3 & 16.7 & 16.7 & 24.5 & 26.4 \\
MAC-SQL & 10.8 & \phantom{0}8.1 & 27.6 & 20.7 & 23.5 & 23.5 & \phantom{0}6.7 & 20.0 & \textbf{25.0} & \textbf{25.0} & 18.2 & 17.3 \\
SQLFixAgent & 24.3 & 10.8 & 41.4 & 41.4 & 41.2 & 47.1 & 40.0 & \textbf{66.7} & 16.7 & 16.7 & 32.7 & 32.7 \\
\midrule \addlinespace[0.3em]
\family{13}{Taxonomy-guided refinement} \\
\addlinespace[0.3em]
ErrorLLM & -- & 16.2 & -- & 44.8 & -- & 41.2 & -- & 60.0 & -- & 16.7 & -- & 33.6 \\
\midrule \addlinespace[0.3em]
\family{13}{Taxonomy-guided error grounding} \\
\addlinespace[0.3em]
\sname (ours) & -- & \textbf{37.8} & -- & 51.7 & -- & \textbf{58.8} & -- & \textbf{66.7} & -- & \textbf{25.0} & -- & \textbf{47.3} \\
\bottomrule
\end{tabular}
\end{table}

\clearpage
\section{Additional Results on End-to-end Correction}
\label{app:predicted-types}

This section extends Section~\ref{sec:predicted}: how well the predicted error types match the oracle error types, and how single-error correction changes by error category when TEG uses them instead.

\subsection{Accuracy of predicted error types}
\label{app:detector}

We predict error types with one call to Opus~4.8 (\texttt{claude-opus-4-8}) per query, using adaptive thinking and an output limit of 8{,}192 tokens. The prompt provides the question, issued SQL, and schema DDL, together with short definitions of all 31 taxonomy sub-types. It requests a JSON response containing a correctness decision and a list of error category and sub-type pairs; the reference SQL and oracle annotations are not supplied.

We parse the predicted labels, normalize category and sub-type names to the taxonomy, and discard unrecognized pairs. The resulting predictions are saved and shared across all correctors in the predicted-type comparisons.

Table~\ref{tab:detector} evaluates the error types predicted by Opus~4.8 against the oracle error types on all 549 queries. Exact match falls from 52.7 on single-error queries to 8.8 on two-error queries and 3.5 on three or more errors. Ignoring sub-types raises overall exact match from 16.2 to 20.0.
Opus~4.8 predicts 726 errors against 1{,}160 oracle annotations, under-counting on 58.7\% of queries and over-counting on 6.2\%. It predicts no error on 16.6\% of all queries. On the evaluated single-error subset, empty predictions trigger TEG's unmasked fallback. For nonempty predictions, missed error types can omit needed correction steps. These counts describe prediction quality on all 549 queries; predicted-type correction itself is evaluated only on the single-error subset.

\begin{table}[h]
\centering\small
\caption{\textbf{Predicted types match about half of single-error annotations, few multi-error ones.}
    Accuracy of the error types predicted by Opus~4.8 on all 549 queries, by the oracle number of errors (110 / 297 / 142).
    Exact match requires the predicted annotation to equal the oracle one as a multiset of (category, sub-type) pairs; the second row ignores the sub-type.
    Ignoring sub-types raises overall exact match only from 16.2 to 20.0, and the predictions under-count errors on 58.7\% of queries.}
\label{tab:detector}
\begin{tabular}{l|ccc|>{\columncolor{tabtint}}c}
\toprule
& 1 error & 2 errors & 3+ errors & Overall \\
\midrule
Exact match & 52.7 & \phantom{0}8.8 & \phantom{0}3.5 & 16.2 \\
Exact match, category only & 55.5 & 14.8 & \phantom{0}3.5 & 20.0 \\
\bottomrule
\end{tabular}
\end{table}

\subsection{Single-error results by error category with predicted types}
\label{app:predicted-by-category}

Table~\ref{tab:predicted_by_category} breaks down the results with oracle and predicted types by category. The Opus~4.8 block uses only the types it predicts itself.
For TEG, predicted types reduce Attribute accuracy by 13.5 points at 14B and 13.6 at 32B, and reduce Condition accuracy at all three Qwen2.5 scales. Table and Others results vary in direction across scales. Their small sample sizes (15 and 12 queries) limit interpretation of these reversals, and the oracle row is not a strict upper bound for every category.

\begin{table}[h]
\centering\small
\caption{\textbf{Predicted error types lower \sname{}'s Condition accuracy at every Qwen2.5 scale.}
    Single-error execution accuracy by the oracle error category.
    Each model row is Base, a single unmasked correction call, and the indented rows add \sname{} with predicted or oracle types.
    The Opus~4.8 block uses only the types it predicts itself, and Qwen3-8B has separate thinking OFF/ON blocks.
    Bold marks the best overall accuracy within each model block.
    Attribute accuracy also drops at 14B and 32B, whereas Table and Others change in both directions on only 15 and 12 queries.}
\label{tab:predicted_by_category}
\setlength{\tabcolsep}{4.5pt}
\begin{tabular}{l|ccccc|c}
\toprule
Method & Attribute & Value & Condition & Table & Others & Overall \\
\midrule
\textbf{Qwen2.5-7B} & 24.3 & 41.4 & 29.4 & 33.3 & 16.7 & 30.0 \\
\midrule
\quad \sname (predicted types) & 37.8 & 55.2 & 52.9 & 53.3 & 25.0 & 45.5 \\
\rowcolor{tabtint} \quad \sname (oracle types) & 37.8 & 51.7 & 58.8 & 66.7 & 25.0 & \textbf{47.3} \\
\midrule
\textbf{Qwen2.5-14B} & 27.0 & 37.9 & 41.2 & 40.0 & 41.7 & 35.5 \\
\midrule
\quad \sname (predicted types) & 35.1 & 58.6 & 47.1 & 80.0 & 25.0 & 48.2 \\
\rowcolor{tabtint} \quad \sname (oracle types) & 48.6 & 55.2 & 58.8 & 66.7 & 33.3 & \textbf{52.7} \\
\midrule
\textbf{Qwen2.5-32B} & 32.4 & 48.3 & 35.3 & 66.7 & 33.3 & 41.8 \\
\midrule
\quad \sname (predicted types) & 43.2 & 58.6 & 47.1 & 86.7 & 33.3 & 52.7 \\
\rowcolor{tabtint} \quad \sname (oracle types) & 56.8 & 55.2 & 52.9 & 93.3 & 25.0 & \textbf{57.3} \\
\midrule
\textbf{Qwen3-8B (Thinking OFF)} & 27.0 & 34.5 & 29.4 & 46.7 & 25.0 & 31.8 \\
\midrule
\quad \sname (predicted types) & 37.8 & 51.7 & 35.3 & 80.0 & 25.0 & 45.5 \\
\rowcolor{tabtint} \quad \sname (oracle types) & 43.2 & 48.3 & 41.2 & 80.0 & 16.7 & \textbf{46.4} \\
\midrule
\textbf{Qwen3-8B (Thinking ON)} & 37.8 & 44.8 & 35.3 & 73.3 & 33.3 & 43.6 \\
\midrule
\quad \sname (predicted types) & 54.1 & 51.7 & 52.9 & 73.3 & 33.3 & 53.6 \\
\rowcolor{tabtint} \quad \sname (oracle types) & 62.2 & 48.3 & 58.8 & 73.3 & 41.7 & \textbf{57.3} \\
\midrule
\textbf{Opus~4.8} & 51.4 & 69.0 & 58.8 & 86.7 & 66.7 & 63.6 \\
\midrule
\rowcolor{tabtint} \quad \sname (self-predicted types) & 54.1 & 82.8 & 52.9 & 93.3 & 75.0 & \textbf{69.1} \\
\bottomrule
\end{tabular}
\end{table}

\clearpage
\section{Extended Discussion}
\label{app:discussion}
This section expands the discussion in Section~\ref{sec:conclusion}: what the results say about using a diagnosis, where the idea of error grounding comes from, and what carrying it beyond SQL would require.

\subsection{Using error diagnoses to construct correction inputs}
\label{app:discussion-input}
The comparisons with baselines given the same error types and the masking ablation support using diagnoses to construct correction inputs: TEG retains its lead, and the w/o masking variant lowers accuracy while retaining the types. These comparisons evaluate the full pipelines and editing configurations rather than isolating masking alone.
The 7B gains suggest potential for locally hosted correction in organizations that keep sensitive database context on premises, provided error annotations can also be supplied locally.
The size of the benefit depends on the LLM and on whether the types are oracle or predicted.
The near tie with ErrorLLM under oracle types on Opus~4.8 suggests that a strong LLM also makes good use of the type-specific context that ErrorLLM extracts.

\subsection{Relation to visual grounding and inpainting}
\label{app:discussion-analogy}
Visual grounding links linguistic descriptions to image regions~\citep{yu2018mattnet}, offering an analogy for connecting an error diagnosis to code structures.
TEG grounds types in construct classes and edit instructions, so its targets can include correct occurrences rather than an exact faulty location.
Its masking is inspired by image inpainting~\citep{yu2018generative} and mask-guided diffusion editing~\citep{avrahami2022blended}, where selected content is regenerated from its surroundings.
TEG adopts this regeneration with an LLM that generates complete queries, without diffusion or a fixed edit boundary, so the LLM may also change unmasked constructs.

\subsection{Carrying error grounding beyond SQL}
\label{app:discussion-transfer}
A transfer needs recurring error categories, identifiable editable elements, and instructions that preserve the surrounding constraints.
An AST can expose these elements in code, while a schema can do so for configuration data.
Candidate selection could use tests, type checks, or schema validation where available, although these checks would not by themselves establish semantic correctness.
Sequential correction would also need to account for dependencies between edits and the possibility that one correction invalidates another.
Each domain would keep the mapping from error types to editable structures and requested operations.
In general-purpose code, an API-misuse category could select call expressions and their arguments for replacement.
In configuration files, an incorrect-value category could select fields for replacement, while a missing-key category has no existing construct to select and would request insertion.
These applications are prospective; each domain would need its own masking rules and correction order.
The results here establish evidence in SQL, leaving the effectiveness and cost of these extensions to future evaluation.

\end{document}